\pdfoutput=1

\documentclass[11pt,a4paper]{article}
\usepackage[hyperref]{acl2023}
\usepackage{times}
\usepackage{inconsolata}
\usepackage{latexsym}
\usepackage{danudefs}
\usepackage{algorithmic}
\usepackage{algorithm}
\usepackage{color}
\usepackage{booktabs}
\usepackage{xspace}
\usepackage{url}
\usepackage{braket}
\usepackage{multirow}
\usepackage{graphicx}
\usepackage{microtype}
\usepackage{makecell} 
\usepackage{amssymb}
\usepackage{tikz}
\usepackage{subcaption} 

\usepackage[most]{tcolorbox}
\newtcbox{\greenboxtext}{on line,colback=green!10,colframe=white,size=fbox,arc=3pt,boxrule=0pt}
\newtcbox{\redboxtext}{on line,colback=red!10,colframe=white,size=fbox,arc=3pt,boxrule=0pt}
\newtcbox{\blueboxtext}{on line,colback=green!10,colframe=white,size=fbox,arc=3pt,boxrule=0pt}
\usepackage[english]{babel}
\usepackage{hyperref}
\addto\captionsenglish{%
}
\addto\extrasenglish{%
}

\makeatletter
\newif\if@in@acrolist
\AtBeginEnvironment{acronym}{\@in@acrolisttrue}
\newrobustcmd{\LU}[2]{\if@in@acrolist#1\else#2\fi}
\newcommand{\ACF}[1]{{\@in@acrolisttrue\acf{#1}}}
\makeatother
\usepackage{acronym}
\acrodef{MLM}[MLM]{Masked Language Model}
\acrodef{SoTA}[SoTA]{state-of-the-art}
\acrodef{LLM}[LLM]{Large Language Model}
\acrodefplural{LLM}[LLMs]{Large Language Models}
\acrodef{GCR}[GCR]{Generative Commonsense Reasoning}
\acrodef{ICL}[ICL]{\LU{I}{i}n-context \LU{L}{l}earning}
\acrodef{MoE}[MoE]{Mixture of Experts}
\acrodef{NLG}[NLG]{Natural Language Generation}
\acrodef{NLP}[NLP]{Natural Language Processing}
\acrodef{Prop}[ICD]{In-Context Diversification}
\acrodef{FBD}[FBD]{Fr{\'e}chet BERT Distance}
\acrodef{KG}[KG]{knowledge graph}

\newcommand{\default}{\textsf{default}}
\newcommand{\diversified}{\textsf{diversified}}

\usepackage{todonotes}

\makeatletter
\if@todonotes@disabled

\else

\fi
\makeatother

\title{Improving Diversity of Commonsense Generation by \\Large Language Models via In-Context Learning}

\author{Tianhui Zhang$^\dagger$ \And
    Bei Peng$^\dagger$  \\
  University of Liverpool$^\dagger$, Amazon$^\diamondsuit$\\
   {\tt \{tianhui.zhang, danushka, bei.peng\}@liverpool.ac.uk} \\ \And
  Danushka Bollegala$^{\dagger,\diamondsuit}$}

\date{}

\begin{document}
\maketitle

\begin{abstract}
\ac{GCR} requires a model to reason about a situation using commonsense knowledge, while generating coherent sentences. 
Although the quality of the generated sentences is crucial, the diversity of the generation is equally important because it reflects the model's ability to use a range of commonsense knowledge facts. 
\acp{LLM} have shown proficiency in enhancing the generation quality across various tasks through \ac{ICL} using given examples without the need for any fine-tuning.  
However, the diversity aspect in \ac{LLM} outputs has not been systematically studied before. 
To address this, we propose a simple method that diversifies the \ac{LLM} generations, while preserving their quality. 
Experimental results on three benchmark \ac{GCR} datasets show that our method achieves an ideal balance between the quality and diversity. 
Moreover, the sentences generated by our proposed method can be used as training data to improve diversity in existing commonsense generators.\footnote{The code is available at \url{https://github.com/AvataGarde/In_Context_Diversification}}
\end{abstract}

\section{Introduction}
\label{sec:intro}

Commonsense reasoning is the ability to make logical deductions about concepts  encountered in daily life, and is considered as a critical property of intelligent agents~\cite{Davis_2015}.
Concepts are mental representations of classes and are expressed using words in a language~\cite{liu:2023:dimongen}.
Given the inputs, the \ac{GCR} task requires a model to generate a coherent sentence that is grammatical and adheres to commonsense, evaluated by its similarity to a set of human-written reference sentences covering the same set of concepts~\cite{CommonGen}.

\begin{figure}[t]
\centering
\includegraphics[width=1.0\linewidth]{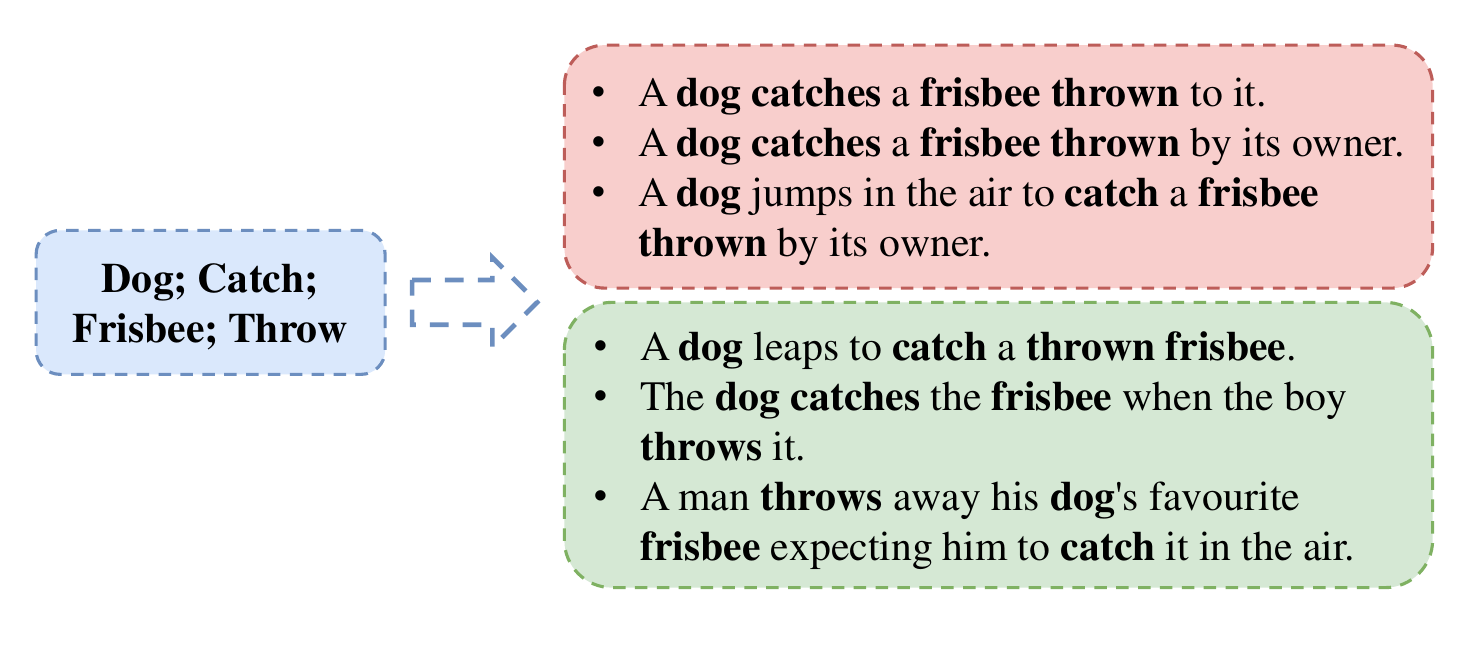}
\caption{An example of diverse generated sentence sets in CommonGen~\citep{CommonGen} dataset. The generation shown at the bottom (in \greenboxtext{green}) is considered by human annotators to be more diverse than those at the top (in \redboxtext{red}).}
\label{fig:commongen}
\end{figure}

Often there exists multiple relationships between a given set of concepts, leading to alternative reasoning paths that take \emph{diverse} view points.
For example, given the four concepts \emph{dog}, \emph{frisbee}, \emph{throw} and \emph{catch}, different sentences can be generated as shown in~\autoref{fig:commongen}.
Although all sentences shown in \autoref{fig:commongen} are grammatical, the bottom set expresses diverse view points (e.g. from the dog's as well as the man's) compared to the set at the top.
Apart from the generation quality, diversity is also an important factor in text generation because the low-diversity texts tend to be dull, repetitive or biased towards a particular view point~\citep{tevet:2021:evaluating}. 
Diversity is an important consideration in many \ac{NLG} applications, such as story generation~\citep{li:2018:story}, paraphrase generation~\citep{gupta:2018:deep}, and \ac{GCR}~\citep{yu:2022:diversifying, liu:2023:dimongen}. 
In \ac{GCR} tasks, diversity requires model's ability to generate explanations for everyday scenarios from various perspectives and to reflect diverse relationships between input concepts. Moreover, \ac{GCR} datasets often contain input texts with limited information. Diversifying \ac{GCR} requires a deep understanding of relationships and commonsense knowledge around the input concepts.
Existing methods promote diversity through special decoding strategies, such as nucleus sampling~\citep{holtzman:2019:sample}, or encoding interventions such as random noise injection~\citep{gupta:2018:deep} or \ac{MoE} approaches~\citep{shen:2019:mixture}.
Temperature sampling is one of the common methods to increase the \ac{LLM} generation diversity. However, it could affect the quality of the generation and lead to nonsensical sequences \citep{peeperkorn:2024:temperature}.

We propose \ac{Prop}, a computationally-efficient and accurate method to improve the diversity in \ac{GCR}, where the sentences are generated from a pre-trained \ac{LLM}, and strikes a fine-balance between the output diversity and quality.  
\ac{Prop} uses an \ac{ICL} approach to increase the diversity of the sentences generated by an \ac{LLM}, while maintaining the quality of the generation.
\ac{Prop} is a two-step process where it first lets an \ac{LLM} to generate sentences that are grammatical, commonsense bearing and cover the tasks' requirements. 
If the diversity is low, \ac{Prop} provides feedback to the \ac{LLM}, instructing it to generate more diverse sentences considering the already generated sentences.
Next, \ac{Prop} uses a diversity-based sampling method to make a trade-off between quality and diversity with a user-specific diversity metric.

Given that \ac{Prop} is using \ac{LLM}s to generate diverse sentences via \ac{ICL} and without updating the parameters of the \ac{LLM}s, an interesting and open question is \emph{whether an \ac{LLM} can accurately judge the diversity of a given set of sentences}.
To answer this question, we conduct an experiment where we instruct \texttt{\texttt{GPT3.5-turbo}} to judge the diversity of the set of input sentences according to a five-scale grading system, and convert the predicted grades into binary judgements (i.e. diverse vs. non-diverse).
We compare the \ac{LLM}-assigned grades against those by a group of human annotators, and find a 
moderate-level (Cohen's Kappa of 0.409) agreement between human vs. \ac{LLM} judgements, demonstrating that \ac{LLM}s can indeed be instructed to obtain diversity judgements for \ac{GCR} tasks.

We evaluate \ac{Prop} on three \ac{GCR} tasks/datasets: CommonGen~\cite{CommonGen}, ComVE~\cite{wang:2020:semeval}, and DimonGen~\cite{liu:2023:dimongen}. 
We find that our proposed \ac{Prop} balances diversity and quality appropriately, improving their harmonic mean by at least 6\% over that of a \default~baseline. 
Moreover, the sentences generated by \ac{Prop} can be used as training data to improve diversity in a Seq2Seq model~\cite{sutskever:2014:sequence,BART:2019}, producing results that are comparable to the models that are trained on knowledge graphs or human-written text corpora~\cite{kgbart:2021,Ekibart:2020,KFCNet:2021}.

\section{Related Work}
\label{sec:related}

\paragraph{Diverse Text Generation.}
 A variety of methods have been proposed to enhance the diversity of \ac{NLG}. 
 Sampling-based decoding is an effective method to increase the generation diversity. 
 \citet{holtzman:2019:sample} proposed nucleus sampling to generate diverse content at the generation stage. 
 Truncated sampling~\citep{fan:2018:hierarchical} prunes and then samples the tokens based on the probability distribution. 
 Furthermore, ~\citet{shen:2019:mixture} proposed an \ac{MoE} approach to diversify translation outputs. 
 Moreover, incorporating external corpora in the \ac{MoE} further promotes diversity, such as by using a knowledge graph~\citep{yu:2022:diversifying,hwang:2023:knowledge} or by a collection of retrieved sentences~\citep{liu:2023:dimongen}. 
 Although \ac{LLM}s have reported superior performance in numerous NLPtasks~\citep{touvron:2023:llama,openai:2023:chatgpt,openai:2023:gpt4}, to the best of our knowledge, diversifying their generations in commonsense reasoning with \ac{ICL} has not been explored in prior work on \ac{GCR}.

\paragraph{In-Context Learning.}
Recent studies demonstrate that \ac{LLM}s can exhibit robust few-shot performance on a variety of downstream tasks through \ac{ICL}~\citep{GPT3:2020}. 
\ac{ICL} is a technique for instructing an \ac{LLM} using one or more examples for a particular text generation task. 
The generated text is conditioned on both the input as well as the instruction prompt.
\citet{wang-etal-2023-label}~show that in \ac{ICL}, label words in the demonstration examples function as anchors, which aggregate semantic information to their word representations in the shallow (closer to the input) layers, while providing that information to the final predictions performed by the deeper (closer to the output) layers.
In contrast to fine-tuning-based methods, \ac{ICL} is computationally lightweight because it does not update the parameters of the \ac{LLM}.
Therefore, \ac{ICL} is an attractive method when integrating task-specific knowledge to an \ac{LLM} by simply changing the prompt and the few-shot examples~\citep{dong:2022:icl}.

\begin{figure*}[t]
\centering
\includegraphics[width=0.76\linewidth]{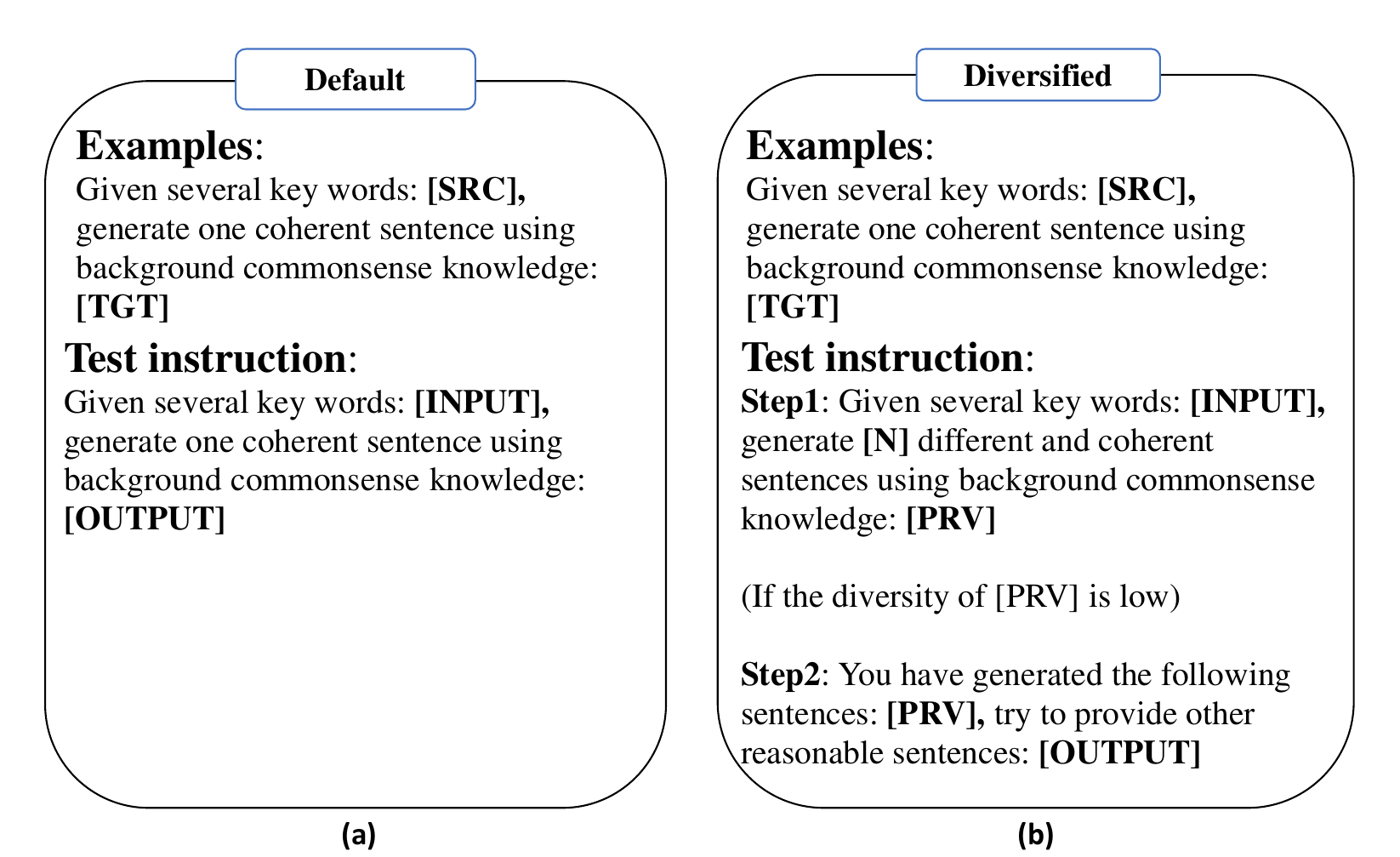}
\caption{An example of \default~and \diversified~prompts is shown for an instance selected from the CommonGen dataset. Here, the \default~prompt shown in \autoref{fig:commongen_prompt}a is taken from~\citet{li:2023:deliberate}. 
Few-shot examples are included in each prompt where \textbf{[SRC]} denotes the set of input concepts and  \textbf{[TGT]} the corresponding sentences in CommonGen.
For a given set of \textbf{[INPUT]} concepts, the \ac{LLM} is then required to generate sentences at the slot \textbf{[OUTPUT]}.
As shown in \autoref{fig:commongen_prompt}b, \ac{Prop} uses the \diversified~prompt, which operates in two steps.
Step 1 generates a set of \textbf{[N]} sentences, \textbf{[PRV]}.
We check for the diversity among the sentences in \textbf{[PRV]}, and if it is low, we use the prompt in Step 2 to generate the final set of sentences.}
\label{fig:commongen_prompt}
\end{figure*}
 
\section{In-context Diversification}
\label{sec:task}

We consider the problem of generating a set of diverse sentences that express commonsense reasoning, either by covering a set of given concepts (in CommonGen and DimonGen) or by providing an explanation for a given counterfactual statement (in ComVE). 
Formally, given a sequence (a set of concepts or a statement) $\cX = \{x_1, \ldots, x_m\}$, the goal of \ac{GCR} is to generate a set of  grammatically correct and commonsense bearing sentences 
$\cY= \{y_1,  \ldots, y_n\}$, where $y_i$ is the $i$-th output generated by the model with probability $p(y_i|\cX)$. 
Moreover, we require that the generated sentences $\{y_1,  \ldots, y_n\}$ to be  lexically as well as semantically diverse.

\subsection{Sentence Generation}
\label{sec:sample_generation}

To explain our proposed \ac{Prop}, let us consider \ac{GCR} on CommonGen, where the models should generate a set of commmonsense bearing sentences $\cY$, such that each sentence contains \emph{all} of the input concepts $\cX$ as shown in \autoref{fig:commongen_prompt}a.
Given an \ac{LLM}, we can design a prompt that contains a task-specific instruction and one or more examples containing the input concepts (denoted by \textbf{[SRC]} in \autoref{fig:commongen_prompt}) and the corresponding human-written sentences containing \textbf{all} given input concepts (denoted by \textbf{[TGT]}) to instruct the \ac{LLM} to generate output sentences $\cY$ (denoted by \textbf{[OUTPUT]}) for a given set of input concepts $\cX$ (denoted by \textbf{[INPUT]}).
We refer to a prompt of this nature as a \default~prompt, and the corresponding set of generated sentences by $\cS_{\rm def}$.

Note that the \default~prompt does not guarantee that the generated set of sentences will be diverse and the \ac{LLM} may return sentences that are redundant.
To address this issue, we propose a \diversified~prompt as shown in \autoref{fig:commongen_prompt}b.

Specifically, the \diversified~prompt operates in two steps.
In Step~1, we require that the \ac{LLM} generate $N$ sentences that are \emph{different}, in addition to being coherent and commonsense bearing.
Next, we use a suitable diversity metric to evaluate the level of diversity among the generated set of sentences.
If the diversity of the generated sentences is low, in Step~2, the sentences would be sent back to the \ac{LLM} and instruct it to generate sentences that are \emph{different} to those.
As the criterion  for triggering Step~2, we check whether the exact same sentence has been generated multiple times by the \ac{LLM} during Step~1.
The final set of generated sentences is denoted by $\cS_{\rm div}$.

\subsection{Diversity-based Sampling} 
\label{sec:sampling}

\begin{algorithm}[t]
    \small
    \algsetup{linenosize=\small}
    \caption{In-Context Diversification (ICD)}
    \label{algo:sampling}
    \begin{algorithmic}
    \renewcommand{\algorithmicrequire}{\textbf{Input:}}
    \renewcommand{\algorithmicensure}{\textbf{Output:}}
    \REQUIRE Generated sets of sentences $\cS_{\rm def}$ and $\cS_{\rm div}$, respectively from \default~and \diversified~prompts, the number of desired output sentences $N$, the temporarily stored value $\alpha$ and a diversity metric $f$.
    \ENSURE Output set of sentences $\cS^{*}$
    \STATE $\cS^{*} \leftarrow \emptyset$
    \STATE $~\alpha~\leftarrow 0$
    \FOR {$\cS \in (\cS_{\rm def} \cup \cS_{\rm div})$}
        \IF {$(|\cS| == N) \land (f(\cS) \geq \alpha)$}
            \STATE $\alpha \leftarrow f(\cS)$
            \STATE $\cS^* \leftarrow \cS$
        \ENDIF
    \ENDFOR
    \RETURN $\cS^{*}$
    \end{algorithmic}
\end{algorithm}

Because of the limited availability of human-written reference sentences for evaluating \ac{GCR} models, there exists a trade-off between quality vs. diversity for \ac{GCR} tasks.\footnote{This trade-off is further empirically verified in \autoref{sec:exp:GCR}.}
Simply maximising for diversity often leads to generations that do not cover the input concepts in a natural way.
For example, a randomly selected set of sentences would be highly diverse, yet unlikely to capture the input concept sets.
On the other hand, if we force an \ac{LLM} to generate sentences that contain all of the input concepts, it might find difficult to generate semantically diverse sentences and resort to trivial lexical or syntactic diversity tricks such as morphological inflections or word-order permutations. 

To address this issue, we propose a diversity-based sampling method shown in~\autoref{algo:sampling}.
Consider that the \default~prompt provides a set $\cS_{\rm def}$ of sentences that have not been optimised for diversity (likely to have a higher quality), while on the other hand the \diversified~prompt provides a set $\cS_{\rm div}$ of sentences that are further refined for diversity (likely to have a higher diversity).
We wish to find a set of sentences that simultaneously satisfies the following criteria:
(a) must contain exactly $N$ sentences, as specified by the user, and
(b) must have a high diversity score, measured using a user-specified diversity metric $f (\in \R_{\geq 0})$.
We formalise this as a subset search problem, where we compute the union $\cS_{\rm def} \cup \cS_{\rm div}$ and search for the subset $\cS^{*}$ that jointly satisfies those criteria following the procedure detailed in \autoref{algo:sampling}.
Although the total number of subsets of size $N$ is ${|\cS_{\rm def} \cup \cS_{\rm div}|} \choose {N}$, it is sufficiently small for the values of $N (\leq 6)$ in our \ac{GCR} tasks, which makes this subset search fast in practice.

\section{Experimental Settings}
\label{sec:exp}

\subsection{Tasks and Datasets}
We evaluate \ac{Prop} on three \ac{GCR} tasks as follows.

\noindent\textbf{Constrained Commonsense Reasoning:}
In CommonGen~\citep{CommonGen} benchmark, a model is required to generate a sentence covering a given set of concepts such that background commonsense knowledge associated with the input concepts is reflected. 
This dataset contains 35K distinct concept sets (train = 32651, dev = 993, and test = 1497) with corresponding human written sentences (train = 67389, dev = 4018, and test = 6042). 
Each instance contains on average 3-5 input concepts.

\noindent\textbf{Commonsense Explanation Reasoning:} ComVE~\citep{wang:2020:semeval} is part of the SemEval 2020 commonsense validation task, where for a given counterfactual statement, a model is required to generate an explanation providing a reason describing why the statement is nonsensical. 
This dataset contains 10K (train = 8532, dev = 476, and test = 992) examples, where each example contains three reference outputs.

\noindent\textbf{Diversified \ac{GCR}:} DimonGen~\cite{liu:2023:dimongen} involves generating diverse sentences that describe the relationships between two given concepts.
It is a challenging task because it requires generating reasonable scenarios for a given pair of concepts without any context.
This dataset contains 17109 instances (train = 15263, dev = 665, test = 1181), where each instance has 3-5 references.

\subsection{Evaluation Metrics}
\label{sec:metrics}

We measure both the quality  and diversity  of the sentences generated by models using the metrics described next.

\subsubsection{Quality Metrics}
\label{sec:quality-metrics}

We compare a generated sentence by a model against a set of human-written references to evaluate the quality of the generation using several metrics:
BLEU~\citep{bleu} measures $n$-gram precision against human reference texts, 
SPICE~\citep{spice} measures the semantic propositional overlap between two sentences, and
BERTScore~\citep{zhang:2020:bertscore} uses contextualised word embeddings to measure the semantic similarity between tokens in two sentences.  
In alignment with prior works~\citep{yu:2022:diversifying,liu:2023:dimongen, hwang:2023:knowledge}, when multiple candidate sentences are generated for a test case, we select the highest-scoring candidate for evaluating quality.

\subsubsection{Diversity Metrics}
\label{sec:diversity-metrics}

\paragraph{Pairwise Diversity:} 
We use self-BLEU~\citep{zhu:2018:selfbleu} to measure $n$-gram overlap among sentences within each generated set. 
The metric computes the average sentence-level similarity between all pairwise combinations of the generations in the generation set. 
Note that unlike BLEU, self-BLEU does \emph{not} require human generated references for measuring diversity.
We use self-BLEU3/4 (corresponding to $n = 3$ and $4$) in our experiment. 
Lower self-BLEU scores indicate higher lexical diversity.

\paragraph{Corpus Diversity:}
To measure the variety within our generated text corpus, we employ Distinct-$k$~\citep{li:2016:distinct}, which calculates the ratio of unique $k$-grams to the total number of $k$-grams. 
This metric is particularly useful for adjusting the bias of LLMs toward generating longer sequences, ensuring that diversity is not artificially inflated by the sentence length. 
Additionally, we use Entropy-$k$ to evaluate the distributional uniformity of $k$-gram occurrences, considering word frequencies for a more nuanced view of diversity.
Higher Distinct-$k$ and Entropy-$k$ scores indicate higher diversity.

\paragraph{Semantic Diversity:} 
All previously described diversity metrics are limited to evaluating lexical diversity.
To measure diversity at a semantic level, we propose self-cosSim, which is the average pairwise cosine similarity between generated sentences, computed using sentence embeddings obtained from SimCSE~\citep{gao:2021:simcse}.
Likewise, we define the self-BERTScore as a diversity metric that averages the BERTScores for all generated sentence pairs.
Lower self-cosSim and self-BERTScore values indicate  higher semantic diversity.

\subsubsection{Combined Metrics} 
\label{sec:combined-metrics}

We would prefer \ac{GCR} models that have both high quality and high diversity.
To incoporate both aspects into a single metric, we compute the \textbf{Harmonic Mean} between 
(a) the self-BLEU-4 as the diversity metric, and
(b) BERTScore as the quality metric.
As discussed in~\autoref{sec:sampling}, there exists a trade-off between quality and diversity in \ac{GCR}.
Therefore, the harmonic mean is suitable when averaging quality and diversity scores.\footnote{We use self-BLEU-4 for diversity and BERTScore for quality  in Harmonic Mean due to their reliability shown in preliminary evaluations. Other metric pairs are in~\autoref{sec:harmonic}. }

\citet{FBD}~proposed \ac{FBD} as a joint metric for simultaneously measuring both the quality and diversity of \ac{NLG}.
\ac{FBD} is inspired by the Fr{\'e}chet Inception Distance (FID), proposed by \citet{FID}, for measuring the quality of image generation.
Specifically, \ac{FBD} computes the pooler output\footnote{The last layer's hidden-state of the first token of the sequence is further processed by a Linear layer and a Tanh activation function.} of a sentence as its embedding~\cite{BERT} and represents a set of sentences using the mean vector and the covariance matrix computed from their sentence embeddings.
Next, Wasserstein-2 distance is computed between the set of reference sentences and the set of generated sentences, which captures both the distance between the means as well as variance in the distributions.
Lower \ac{FBD} scores indicate high combined performance.

\subsection{Implementation Details}
\label{sec:implementation}
We use \texttt{GPT3.5-turbo} and \texttt{Vicuna-13b-v1.5}\footnote{\url{https://huggingface.co/lmsys/vicuna-13b-v1.5}} as \ac{LLM}s with temperature set to 1.0 in our experiments.
By using two \ac{LLM}s with significantly differing number of parameters and by including, Vicuna, an open source \ac{LLM}, we plan to improve the reliability and reproducibility of our results.
Max response length is set to 25 tokens. 
The inference times for CommonGen, ComVE and DimonGen datasets are respectively 5-6, 2-3 and 1-2 hours.
The costs of running \ac{Prop} with \texttt{GPT3.5-turbo} are ca. \$6, \$4 and \$4 respectively for CommonGen, ComVE and DimonGen datasets.
On the other hand, the costs of fine-tuning on \texttt{GPT3.5-turbo} are much higher at \$58.8 for CommonGen, \$24.7 for ComVE and \$32.0 for DimonGen.
Moreover, fine-tuning with LoRA~\cite{LORA}  with rank of 8 and alpha of 16 on Vicuna takes ca. 34 hours. 
We use \texttt{BART-large}\footnote{\url{https://huggingface.co/facebook/bart-large}} for \ac{MoE}-based models. 
We use the \texttt{GPT3.5-turbo} to generate sentences for the CommonGen train/dev/test sets using the \default, \diversified~and for \ac{Prop}.
For model training, we use the Adam optimiser~\cite{ADAM} with a batch size of 64, a learning rate of 3e-5 and a beam size of 5. 
All of the \ac{MoE}-based models are trained for 20 epochs and required to generate $k = 3$ sentences.
All experiments, except with GPT3.5-turbo, are conducted on a single RTX A6000 GPU.

\section{Results and Discussion}
\label{sec:results}

\subsection{Commonsense Generation}
\label{sec:exp:GCR}

\begin{table*}[t]
\centering
\resizebox{\textwidth}{!}{
\begin{tabular}{@{}lrrrrrrrrrrrr@{}}
\toprule
                               & \multicolumn{2}{c}{Semantic Diversity $\Downarrow$} & \multicolumn{2}{c}{Corpus Diversity $\Uparrow$} & \multicolumn{2}{c}{Pairwise Diversity $\Downarrow$} & \multicolumn{4}{c}{Quality $\Uparrow$}                                & \multicolumn{2}{c}{Combined} \\ 
                               \cmidrule(r){2-3} \cmidrule(lr){4-5} \cmidrule(lr){6-7} \cmidrule(lr){8-11} \cmidrule(l){12-13}
                               & self-cosSim & self-BERTScore & Entropy-4 & Distinct-4 & self-BLEU-3 & self-BLEU-4 & BLEU-3 & BLEU-4 & SPICE & BERTScore & Harmonic $\Uparrow$& FBD $\Downarrow$\\ 
\midrule
\multicolumn{13}{@{}c}{\textbf{CommonGen}} \\
\midrule
\textbf{Human}                       &67.3    &60.6    &10.9       &91.0        &25.4   &17.6   &-    &-   &-  &-   &-    &- \\
\textbf{Fine-tune}                       & \textit{64.7} & \textit{55.9}  & \textit{11.4}      & \textit{91.1}       & 26.9 & 17.9 & 41.2   & 32.1   & 30.3  & 64.2      & 72.1          & 51.9 \\
\midrule
\textbf{\default}            & 93.3          & 88.7           & 10.2      & 53.7       & 77.2 & 72.4 & \textit{\textbf{50.8}}   & \textit{\textbf{40.9}}   & \textit{\textbf{30.1}}  & \textit{\textbf{70.4}}      & 39.6          & 60.2 \\
\textbf{\diversified}           & 85.2          & 69.8           & \textbf{11.0}      & 83.7       & 44.4 & 34.9 & 44.3   & 34.6   & 28.5  & 65.0      & 65.4          & 53.9 \\
\textbf{\ac{Prop}}            & \textbf{83.5}          & \textbf{66.2}           & \textbf{11.0} & \textbf{88.5} & \textit{\textbf{31.0}} & \textit{\textbf{21.0}} & 47.4   & 37.7   & 29.1  & 67.4      & \textit{\textbf{72.7}}          & \textit{\textbf{51.8}} \\
\midrule
\midrule
\multicolumn{13}{@{}c}{\textbf{ComVE}} \\
\midrule
\textbf{Human}                       &62.7    &47.0    &9.6       &\textit{96.1}        &12.4   &8.1   &-    &-    &-  & -  & -   &- \\
\textbf{Fine-tune}                       & \textit{59.8}   & \textit{42.6}    & \textit{9.8}      & 95.2       & \textit{13.4}  & 10.3  & 27.4   & 19.4   & 33.1 & 53.7     & 67.2          & \textit{47.6} \\
\midrule
\textbf{\default}            & 83.9          & 73.5           & 9.6      & 74.3       & 50.8 & 45.2 & 27.5 & 19.7 & \textit{\textbf{36.2}}  & 55.1  & 54.9      & 50.9 \\
\textbf{\diversified}          & 76.0          & 56.5           & 9.7      & 88.0       & 23.3 & 16.6  & \textit{\textbf{30.6}}   & \textit{\textbf{22.0}}    & 35.8  & \textit{\textbf{56.5}}      & 67.4          & \textbf{47.9} \\
\textbf{\ac{Prop}}           & \textbf{72.5}          & \textbf{51.1}           & \textit{\textbf{9.8}}      & \textbf{90.1} & \textbf{13.7} & \textit{\textbf{8.7}} & 29.0   & 20.8   & 36.1  & 55.5     & \textbf{\textit{69.0}}          & 48.7 \\
\midrule
\midrule
\multicolumn{13}{@{}c}{\textbf{DimonGen}} \\
\midrule
\textbf{Human}                       &56.8    &47.0    &10.1       &85.6        &14.7   &8.7   &-    &-    &-  &-   &-    &- \\
\textbf{Fine-tune}                       &\textit{43.4}    &\textit{33}    & 10.4      &\textit{98.7}        &6.8   &\textit{3.4}   &\textit{17.7}    &\textit{10.7}    &15.5  &42   &\textit{58.5}    &\textit{51.6} \\
\midrule
\textbf{\default}            & 75.7         & 71.3           & 10      & 83.2       & 43.4 & 37.3 & \textbf{15.9}   & \textbf{9.5}   & \textit{\textbf{16.4}}  & \textit{\textbf{44.5}}      & 52.1          & 68.2\\
\textbf{\diversified}           & 57.1          & 46.9           & \textit{\textbf{10.5}}      & 95.9       & 11.2 & 6.5 & 11.4   & 6.4   & 15.2  & 39.9      & 55.9          & 69.0 \\
\textbf{\ac{Prop}}            & \textbf{56.7}          & \textbf{45.7}           & 10.4 & \textbf{96.3} & \textit{\textbf{6.5}} & \textbf{3.5} & 13.2   & 7.6   & 15.4 & 41.7      & \textbf{58.2}          & \textbf{68.0} \\
\bottomrule
\end{tabular}
}
\caption{Diversity and quality scores on CommonGen, ComVE and DimonGen with \texttt{GPT3.5-turbo} \ac{LLM}. 
Best results on each task for each metric are shown in \textit{italics}, while the best performing \ac{ICL} results are shown in \textbf{bold}.
}
\label{tbl:main-res}
\end{table*}

We compare the commonsense generations made by \ac{Prop} against those using the \default~and \diversified~prompts.
For this purpose, we use \texttt{GPT3.5-turbo} as the \ac{LLM} and use the same 10 few-shot examples in all prompts for \ac{ICL}.
Further templates of the \default~and \diversified~prompts used for each task are given in~\autoref{sec:LLMTemplates}.
 To assess the impact of \ac{ICL}, we compare against \textbf{fine-tune} method, wherein \texttt{GPT3.5-turbo} is fine-tuned on the entire training set in each dataset.
Specifically, we use multiple human-written sentences, available in the training data for the three datasets to separately fine-tune the models for each task.
It is noteworthy that the \textbf{fine-tune} method uses a substantially larger dataset for training (e.g., 67,389 sentences from CommonGen) compared to the 10 examples used by the \ac{ICL}-based approaches.
We use self-BLEU-3 as the diversity metric $f$ in \autoref{algo:sampling} for \ac{Prop} in this evaluation.
The outcomes, presented in~\autoref{tbl:main-res}, highlight the diversity and quality metrics of these methods across the CommonGen, ConVE, and DimonGen datasets.
Additionally, a \textbf{human} baseline is introduced to evaluate the diversity of sentences written by humans, where we pair-wise compare the human-written sentences for each input in the instances in the benchmark datasets using diversity metrics.
Note that however, the \textbf{human} baseline must not be considered as an upper-bound for diversity because there are only a smaller number of human-written sentences per instance in the benchmark datasets.

From \autoref{tbl:main-res}, we see that \textbf{fine-tune} generates sentences that have high semantic and corpus diversity, and outperforms the \textbf{human} baseline.
However, recall that \textbf{fine-tune} requires a much larger training set and is computationally costly compared to all \ac{ICL}-based methods. 
Moreover, we see that \ac{Prop} can strike a good balance between quality and diversity in the
sentences generated. 
Among the \ac{ICL}-based methods, \ac{Prop} achieves the best diversity scores on all diversity metrics in all three datasets. It also exhibits higher diversity compared against the human-written references. 
Moreover, \ac{Prop} outperforms \default~and~\diversified~according to the Combined metrics. \ac{Prop} also achieves a Harmonic Mean comparable to that of the \textbf{fine-tune} baseline. 
Although \default~reports the best quality scores, it has low diversity, and is consistently outperformed by \diversified~and \ac{Prop} on diversity metrics. 
On the other hand, \diversified~generally scores lower on the quality metrics. 
Compared to \default~and \diversified, \ac{Prop} enhances generation diversity while maintaining a satisfactory level of quality. 
ICD is also more stable to the sampling method such as temperature than \textbf{fine-tune}, as shown in~\autoref{sec:temperature}.
Note that \textbf{fine-tune} is not an \ac{ICL} setting (the focus of this paper) and is included only as a baseline to demonstrate the level of performance that can be achieved by fine-tuning on a much larger dataset.
Despite this, \ac{Prop} outperforms \textbf{fine-tune} on the Pairwise Diversity in all three datasets, and Combined metrics in the CommonGen dataset.

\begin{table*}[t]
\centering
\small
\begin{tabular}{@{}lrrrrrrrrr@{}}
\toprule
Method                           & SCS $\Downarrow$ & SBS $\Downarrow$ & E-4$\Uparrow$ & D-4$\Uparrow$ & SB-3$\Downarrow$ & BLEU-3$\Uparrow$ & SPICE$\Uparrow $ & HM $\Uparrow$ & FBD $\Downarrow$ \\ 
\midrule
\textbf{Fine-tune}                              & 59.6           & 49.9        & \textit{11.4} & 93.3  & 22.8   & 35.8 & 27.6 & 69.9 & 52.4 \\
\midrule
\textbf{Default}                              & 82.2           & 73.8        & 10.9 & 74.9   & 52.9   & \textit{\textbf{44.6}} & \textit{\textbf{29.1}} & 60.2 & 56.2 \\
\textbf{Diversified}                           & \textit{\textbf{59.1}}           & 53.3    & 11.3 & 91.3       & 23.6 & 32.6 & 24.3   & 68.6   & 53.2 \\
\textbf{\ac{Prop}}                      & 59.3           & \textit{\textbf{49.8}}     & \textbf{11.3} & \textit{\textbf{93.7}}     & \textit{\textbf{11.3}} & 34.2 & 25.5 & \textit{\textbf{73.4}}  & \textit{\textbf{51.0}} \\

\bottomrule
\end{tabular}
\caption{\ac{GCR} on CommonGen using Vicuna-13b. \ac{Prop} uses self-BLEU-3. Here, SCS: self-CosSim, SBS: self-BERTScore, E-4: Entropy-4, D-4: Distinct-4, SB-3: self-BLEU3, HM: Harmonic Mean. Best results for each metric are shown in \textit{italics}, while the best performing \ac{ICL} results are shown in \textbf{bold}.}
\label{tbl:vicuna-commongen}
\end{table*}

As an open source alternative \ac{LLM} to \texttt{GPT3.5-turbo}, we repeat this evaluation with \texttt{Vicuna-13b}~\citep{zheng:2023:judging} in \autoref{tbl:vicuna-commongen}.
The same 10 few-shot examples as used with \texttt{GPT3.5-turbo} are used in this experiment for the \ac{ICL}-based methods.
Full table on three datasets are shown in~\autoref{sec:full-vicuna}.
\autoref{tbl:vicuna-commongen} reconfirms \ac{Prop}'s ability to balance both quality and diversity according to the Combined metrics (i.e. Harmonic Mean and FBD) on this dataset.
Interestingly, we see that methods that use Vicuna-13b to be more diverse compared to those that use \texttt{GPT3.5-turbo}, while the latter showing better generation quality.

\begin{table*}[t]
\centering
\small
\begin{tabular}{@{}lrrrrrrrrr@{}}
\toprule
Method                           & SCS $\Downarrow$ & SBS $\Downarrow$ & E-4$\Uparrow$ & D-4$\Uparrow$ & SB-3$\Downarrow$ & BLEU-3$\Uparrow$ & SPICE$\Uparrow $ & HM $\Uparrow$ & FBD $\Downarrow$ \\ 
\midrule
\textbf{self-BLEU-3 }                    & 83.5           & 66.2          & \textbf{11.0} & \textbf{88.5} & \textbf{31.0} & 47.4 & 29.1 & \textbf{72.7}   & \textbf{51.8} \\
\textbf{self-CosSim}             & \textbf{81.0}  & 70.1   &10.9  & 82.5         & 44.5  & \textbf{47.6}  & \textbf{29.3}     & 65.7     & \textbf{51.8} \\
\textbf{self-BERTScore}                 & 83.1           & \textbf{62.8}  & \textbf{11.0} & 87.0 & 36.3 & 46.5 & 28.9     & 69.6 & \textbf{51.8} \\

\bottomrule
\end{tabular}
\caption{Comparing the effect of using different diversity metrics, $f$, in \autoref{algo:sampling} for \ac{Prop}. We use \texttt{GPT3.5-turbo} as the \ac{LLM} and the best results on CommonGen dataset are in \textbf{bold}.
Here, SCS: self-CosSim, SBS: self-BERTScore, E-4: Entropy-4, D-4: Distinct-4, SB-3: self-BLEU3, HM: Harmonic Mean.}
\label{tbl:f}
\small{}
\end{table*}

In \autoref{tbl:f}, we use different diversity metrics as $f$  in \autoref{algo:sampling} to study the effect on text generation of \ac{Prop}.
We see that self-BLUE-3 and self-CosSim perform similarly across the quality metrics.
Self-BERTScore shows a slightly lower quality (BLEU-3 and SPICE).
According to the combined metrics, any of those diversity metrics can be used with \ac{Prop} to obtain comparable performance.
\begin{table*}[t]
\centering
\resizebox{\textwidth}{!}{%
\begin{tabular}{@{}lrrrrrrrrrrrr@{}}
\toprule
                               & \multicolumn{2}{c}{Semantic Diversity $\Downarrow$} & \multicolumn{2}{c}{Corpus Diversity $\Uparrow$} & \multicolumn{2}{c}{Pairwise Diversity $\Downarrow$} & \multicolumn{4}{c}{Quality $\Uparrow$} & \multicolumn{2}{c}{Combined} \\ 
                               \cmidrule(r){2-3} \cmidrule(lr){4-5} \cmidrule(lr){6-7} \cmidrule(lr){8-11} \cmidrule(l){12-13}
                               & self-cosSim & self-BERTScore & Entropy-4 & Distinct-4 & self-BLEU-3 & self-BLEU-4 & BLEU-3 & BLEU-4 & SPICE & BERTScore & Harmonic Mean $\Uparrow$ & FBD $\Downarrow$\\ 
\midrule
\textbf{KG-BART}                         & -           & -              & -         & -          & -    & -    & 42.1      & 30.9   & 32.7  & -         & -             & -    \\
\textbf{EKI-BART}                           & -           & -              & -         & -          & -    & -    & 46.0      & 36.1   & 33.4  & -         & -             & -    \\
\textbf{KFCNet-w/o FC}                          & -           & -              & -         & -          & -    & -    & 50.2      & 42.0   & 35.9  & -         & -             & -    \\
\textbf{KFCNet}                           & -           & -              & -         & -          & -    & -    & \textbf{57.3}      & \textbf{51.5}   & \textbf{39.1}  & -         & -             & -    \\
\midrule
\textbf{MoE}                          & 89.3        & 81.9           & 9.7     & 61.6       & 63.1 & 56.6 & 49.0   & 38.5   & 33.5  & 70.6      & 53.8          & 61.7 \\
\textbf{MoKGE}                         & 88.7        & 80.6           & \textbf{9.9}      & \textbf{65.2}       & 60.4 & 53.6 & 48.8   & 38.4   & 33.1  & 70.3      & 55.9          & 60.8 \\
\midrule
\textbf{\default+MoE}                   & 91.2        & 84.6           & 9.7      & 60.3       & 66.5 & 60.0 & 51.2   & 40.6   & \underline{34.8}  & 72.9      & 51.6          & 62.3 \\
\textbf{\diversified+MoE}               & \textbf{86.7}        & \textbf{80.4}           & 9.8      & 63.3       & 59.2 & 53.5 & 50.7   & 40.6   & 34.0  & 71.3      & 56.3             & \textbf{55.0}    \\
\textbf{\ac{Prop}+MoE}                  & 91.1        & 82.6           & 9.8      & 64.8      & \textbf{59.0} & \textbf{51.1} & \underline{52.4} & \underline{42.2} & 34.5 & \textbf{\underline{73.5}} & \textbf{58.7}  & 62.3 \\
\bottomrule
\end{tabular}%
}
\caption{Downstream evaluation of the \ac{LLM}-generated sentences.
Top block methods use human-generated resources for training, while the ones in the bottom block are trained on \ac{LLM}-generated sentences.
\ac{MoE} approaches are shown in the middle block and bottom block. \texttt{BART-large} is used as the generator for MoE-based methods. Best results for each metric are shown in \textbf{bold}, while the best performing MoE for quality is shown in \textbf{underline}.}
\label{tbl:corpus}
\end{table*}

\begin{figure}[t]
\centering
\includegraphics[width=1.0\linewidth]{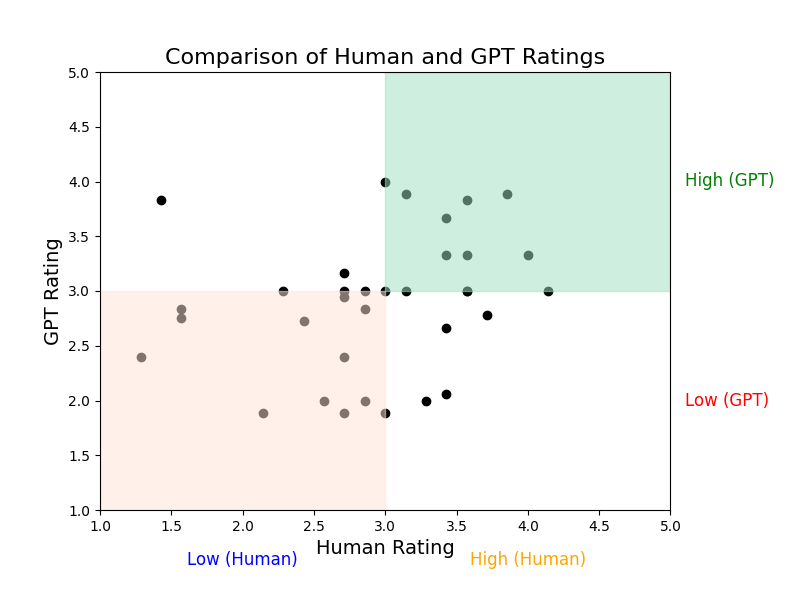}
\caption{Human vs. GPT3.5 diversity ratings for randomly sampled sets of sentences generated by \ac{Prop}. Cohen's $\kappa = 0.409$ indicates a moderate level of agreement.}
\label{fig:human_evaluation}
\end{figure}

\begin{figure*}[t]
\centering
\includegraphics[width=1.0\linewidth]{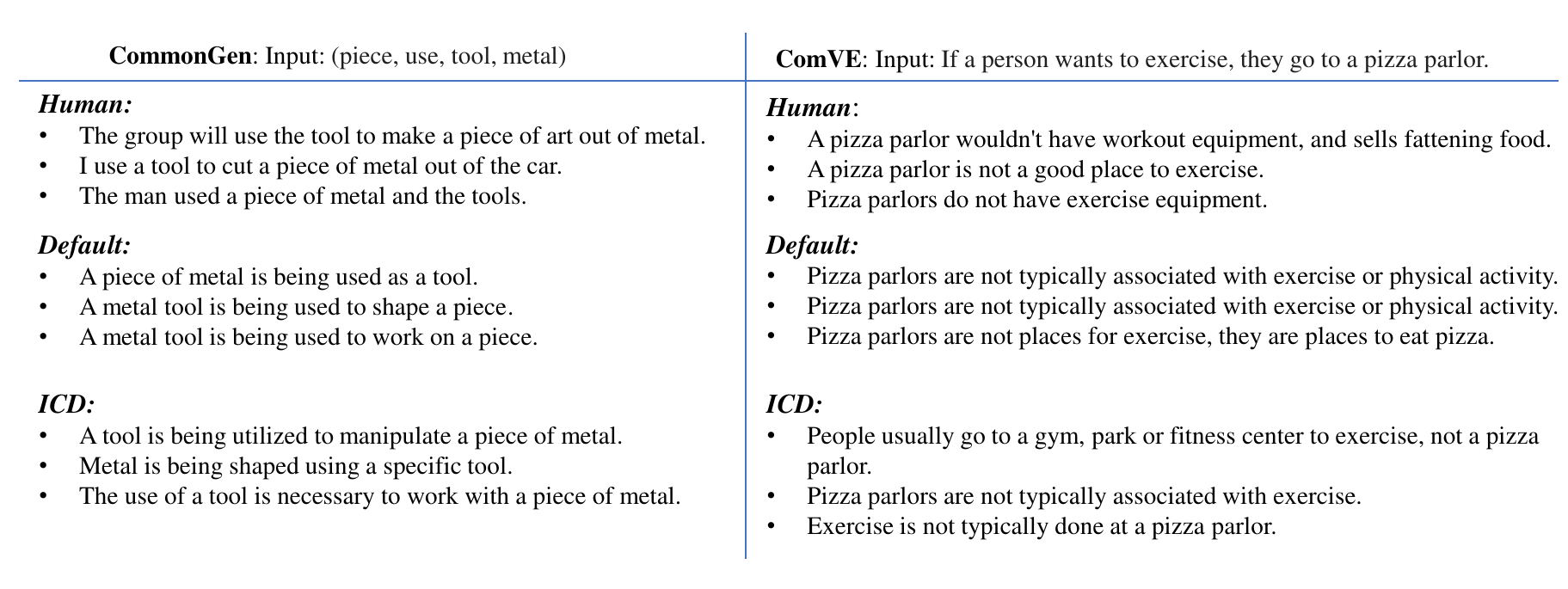}
\caption{Sentences generated by \default~prompt and \ac{Prop} against those by humans on CommonGen and ComVE test instances. \ac{Prop} generates more diverse and high quality sentences than \default.}
\label{fig:sample}
\end{figure*}

\subsection{Downstream Evaluation}
\label{sec:exp:extrinsic}

The experiments presented in \autoref{sec:exp:GCR} show the ability of our proposed \ac{Prop} to generate diverse and commonsense bearing sentences.
Therefore, an important question with practical implications is whether we can use the sentences generated by \ac{Prop} as additional training data to improve both diversity and quality of previously proposed models on the \ac{GCR} task, which could be seen as a downstream (extrinsic) evaluation.

For this purpose we select the \ac{MoE}~\citep{shen:2019:mixture}, which diversifies the generation by selecting outputs from a mixture of experts. 
Each expert is assigned a randomly generated sequence of tokens, which is used as a prefix for all inputs sent to that expert.
For each input, an expert is selected according to the value of a latent variable, which is trained using the hard-EM algorithm.
We follow~\citet{liu:2023:dimongen} and train three experts that retrieve sentences from the collection of sentences generated by \ac{Prop} for concept sets in the CommonGen train split (210846 sentences in total).
We use \texttt{BART-large}~\cite{BART:2019} as the base model, which has shown to produce high quality commonsense generations~\citep{zhang-etal-2023-learning-predict} as the generator for all experts (see \autoref{sec:moe} for further details).
We denote this method by ICD+MoE.

As baselines for comparisons, we repeat the above process using the sentences generated by \default~and \diversified, which we denote respectively as \default+MoE and \diversified+MoE in \autoref{tbl:corpus}.
Moreover, we compare the performance against two previously proposed \ac{MoE} models: \ac{MoE} ~\citep{shen:2019:mixture} and MoKGE~\citep{yu:2022:diversifying}. 
\ac{MoE} relies solely on the base model, whereas MoKGE requires each expert to use different sets of concepts from the ConceptNet~\citep{speer:2017:conceptnet} \ac{KG}. 
Because \citet{yu:2022:diversifying} do not evaluate their MoKGE method on CommonGen, we ran their original implementation\footnote{\url{https://github.com/DM2-ND/MoKGE}} on CommonGen and report results in \autoref{tbl:corpus}.

All previously proposed \ac{GCR} methods are exclusively trained using human-created data (e.g. sentences written by human and/or manually compiled \ac{KG}s such as ConceptNet), whereas the methods described thus far in this section are trained on sentences generated by an \ac{LLM} (\texttt{\texttt{GPT3.5-turbo}}).
Therefore, to evaluate the feasibility of using \ac{LLM}-generated sentences for training \ac{GCR} models, we include the following previously proposed \ac{GCR} models that are trained using a combination of corpora and KGs:
KG-BART~\citep{kgbart:2021},EKI-BART~\citep{Ekibart:2020} and KFCNet~\citep{KFCNet:2021}.
For KFCNet, we present its two results -- KFCNet w/o FC (without Filtering and Contrastive modules), which relies only on sentences including the input concepts, without further processing, and KFCNet, which additionally ranks candidates and adds contrastive modules for the encoder and the decoder~\citep{KFCNet:2021}.
However, note that those methods do \emph{not} consider diversification, and do not report performance using diversity metrics. 
Therefore, we report only their published results for generation quality in \autoref{tbl:corpus}.

From \autoref{tbl:corpus} we see that \diversified+MoE always outperforms the original \ac{MoE} in all diversity metrics, which shows that sentences generated from \ac{LLM}s can be used to diversify \ac{MoE}-based \ac{GCR}.
ICD+MoE closely matches the performance of \diversified+MoE on diversity metrics, while outperforming both \diversified+MoE and \default+MoE on quality metrics.
In particular, the quality metrics reported by ICD+MoE (underlined in \autoref{tbl:corpus}) are competitive against those obtained by the models that are trained on human-compiled resources (in the top block), except against KFCNet.
This finding hints at potential improvement gains for \ac{GCR} by using hybrid training resources that combine both human-compiled and \ac{LLM}-generated data, which we highlight as an interesting future research direction.

\begin{table}[t]
\resizebox{\columnwidth}{!}{%
\begin{tabular}{@{}lllll@{}}
\toprule
\textbf{Dataset \& Metrics} & \textbf{Fine-tune} & \textbf{Default} & \textbf{Diversified} & \textbf{ICD} \\
\midrule
\textbf{CommonGen} \\
self-BLEU4 + BERTScore & 72.1 & 39.6 & 65.4 & \textbf{\textit{72.7}} \\
self-cosSim + SPICE    & \textit{32.6} & 11.0 & 19.5 & \textbf{21.1} \\
self-BERTScore + BLEU3 & \textit{42.6} & 18.5 & 35.9 & \textbf{39.5} \\
\midrule
\textbf{ComVE} \\
self-BLEU4 + BERTScore & 67.2 & 54.9 & 67.4 & \textbf{\textit{69.0}} \\
self-cosSim + SPICE    & \textit{36.3} & 22.3 & 28.7 & \textbf{31.2} \\
self-BERTScore + BLEU3 & \textit{37.1} & 27.0 & 35.9 & \textbf{36.4} \\
\midrule
\textbf{Dimongen} \\
self-BLEU4 + BERTScore & \textit{58.5} & 52.1 & 55.9 & \textbf{58.2} \\
self-cosSim + SPICE    & \textit{24.3} & 19.6 & 22.4 & \textbf{22.7} \\
self-BERTScore + BLEU3 & \textit{28.0} & 20.5 & 18.8 & \textbf{21.2} \\
\bottomrule
\end{tabular}%
}
\caption{Different combined metrics that are calculated as the harmonic means between quality and diversity metric pairs. The metric with `self' in each line is a diversity metric and the other is a quality metric. Best results on each task for each metric are shown in \textit{italics}, while the best performing \ac{ICL} results are shown in \textbf{bold}.}
\label{tab:harmonic-means}
\end{table}

\subsection{Candidate metrics for calculating Harmonic Means}
\label{sec:harmonic}
In the previous experiments, we computed the harmonic mean between self-BLEU-4 and BERTScore to obtain a combined metric that considers both quality and diversity of commonsense generation. 
Specifically, self-BLEU~\citep{zhu:2018:selfbleu} evaluates the $n$-gram overlap between pairs of sentences in the generated set, providing a measure of lexical diversity. 
On the other hand, BERTScore~\citep{zhang:2020:bertscore} assesses the semantic similarity between the generated sentences and the human-written sentences in each dataset, capturing the quality aspects from a semantic perspective.
Note that other combinations of quality and diversity metrics can also be used for computing different harmonic means as shown in \autoref{tab:harmonic-means}.
From \autoref{tab:harmonic-means}, we see that according to each combined metric, ICD achieves the best performance among all ICL-based approaches. 
Moreover, ICD also has comparable performance against the fine-tune method.

\subsection{Diversity-Awareness of \ac{LLM}s}
\label{sec:exp:LLMvsHuman}

Given that we use \ac{LLM}s to produce diverse generations via \ac{ICL}, it remains an open question whether an \ac{LLM} would agree with humans on the diversity of a given set of sentences.
To answer this question, we use randomly selected 210 sentences (35 sets, each containing 6 sentences) generated by \ac{Prop} (using self-BLEU-3 as the diversity metric) for the input concept sets in the CommonGen dataset.
We use \texttt{\texttt{GPT3.5-turbo}} to rate the diversity of a set of sentences according to five levels from 1 (highly similar) to  5 (highly diverse).\footnote{Detailed prompt templates are shown in \autoref{sec:LLMTemplates}.}
We provide the same instructions as the annotation guidelines for eight human annotators, who are graduate students in \ac{NLP}.
To reduce subjective variability in human judgements, we average and then normalise the ratings following the Likert scale.

In \autoref{fig:human_evaluation}, we plot the GPT-assigned ratings against those provided by humans.
We further split the ratings into \emph{high} vs. \emph{low} diversity ratings depending on whether the rating is greater or lesser than 3.
The majority of the data points are distributed along the diagonal quadrants and a Cohen's Kappa of $0.409$ indicating a moderate level of agreement between GPT and human ratings for diversity.

The generated sentences using the \default~prompt, \ac{Prop} and the human references in CommonGen and ComVE datasets for a single test instance are shown in \autoref{fig:sample}.
From \autoref{fig:sample} we see that the sentences generated using the \default~prompt often results in significant token overlap, thereby lowering the diversity.
On the other hand, although \autoref{tbl:main-res} shows that \ac{Prop} has lower scores in quality metrics, we find that  \ac{Prop} generates sentences of sufficient quality. 
These sentences are both lexically and semantically diverse, covering the diverse viewpoints found in the human references. 
We provided additional ICD generated sentences in \autoref{tab:more_exe}, showing that ICD achieves better diversity than the \default~prompt while maintaining adequate quality, and is fluent and commonsense bearing.

\section{Conclusion}
\label{sec:conclusion}
We proposed, \ac{Prop}, an \ac{ICL}-based method for achieving the optimal balance between diversity and quality in text generation via \ac{LLM}s. 
Our experiments, conducted on three \ac{GCR} tasks, demonstrate that \ac{Prop} significantly improves the diversity without substantially compromising the quality. 
Furthermore, we found that by training on the sentences generated by \ac{Prop}, we can improve diversity in previously proposed \ac{GCR} methods.

\section{Limitations}
This study primarily focuses on the generation of English sentences using pre-trained LLMs, a limitation shaped by the datasets we employed. Specifically, we used the ComVE~\citep{wang:2020:semeval}, CommonGen~\citep{CommonGen} and DimonGen~\citep{liu:2023:dimongen} datasets, which are well-regarded for evaluating diversified commonsense reasoning in English. Therefore, our evaluation of the generation quality was limited to English, which is a morphologically limited language. Future research could expand this scope to include multilingual pre-trained models, thereby encompassing a broader linguistic spectrum.
Three GCR datasets contain multiple human-written sentences (3-5 for each sample), which could be considered as golden references for quality evaluation. The focus of our paper is on diversity and in \autoref{sec:exp:LLMvsHuman}, we compare the human and LLM produced ratings for commonsense generation diversity. Therefore, we only conducted additional human evaluation for the diversity of the sentences generated.

Our approach is primarily geared towards optimizing the trade-off between diversity and quality in text generation. Consequently, we maintained consistent default instructions across all experiments, adopting the standard commonsense generation prompts used in~\citet{li:2023:deliberate} as our default instructions.

We note that decoding strategies can influence the diversity of generated outputs. In our work, we fix the decoding strategy (top-p, temperature) to maintain consistency across our experiments. We also evaluate our proposed method under various temperature sampling settings, as shown in \autoref{sec:temperature}. Studying the effect of other decoding strategies on the system could be considered as a future direction.

We conducted our experiments using both a closed model (i.e. \texttt{GPT3.5-turbo-0613}) as well as an open-source one (i.e. \texttt{Vicuna-13b-v1.5}) to promote the reproducibility of our results, which are reported using multiple publicly available benchmarks. 
However, there exist many other \ac{LLM}s with varying numbers of parameters and trained on different corpora.
Therefore, we consider that it is important to evaluate our proposed method on a broad range of \ac{LLM}s to verify the generalisability of our proposed method.
However, conducting such a broad analysis can be computationally intensive and costly.
For example, although GPT-4 is known to have superior text generation capabilities, it incurs substantially higher costs (being 30 times more expensive than \texttt{GPT3.5-turbo} at the current  pricing).
Nevertheless, \ac{Prop} is adaptable and could be extended to other \ac{LLM}s.

\section{Ethical Considerations}
\label{sec:ethics}
The experiments conducted in this paper are based on the publicly available datasets, CommonGen, ComVE, and DimonGen. 
To the best of our knowledge, no ethical issues have been reported for those datasets.
Therefore, we do not foresee any data-related ethical issues arising from our work. 

However, \ac{LLM}s are known to generate responses that may reflect societal biases and potentially harmful content. 
We have not verified whether the \texttt{GPT3.5-turbo} and \texttt{Vicuna-13b} \ac{LLM}s that we use in our experiments have similar problems. 

Therefore, it is important to test on existing benchmarks for social biases and harmful generations before the proposed method is deployed to diversify existing \ac{GCR} methods used by human users.

To elicit human judgements of diversity for the sentences generated by \ac{Prop}, we use annotators who are familiar with working with \ac{LLM}s.
It is possible that their subjective (and possibly biased) viewpoints might have influenced the ratings provided. 
Therefore, it will be important to conduct the evaluation involving a group of annotators with different backgrounds to validate the findings reported in this analysis.

\section*{Acknowledgements}
Danushka Bollegala holds concurrent appointments as a Professor at University of Liverpool and as an Amazon Scholar. This paper describes work performed at the University of Liverpool and is not associated with Amazon.

\bibliography{GCR-order.bib}

\begin{thebibliography}{}
\expandafter\ifx\csname natexlab\endcsname\relax\def\natexlab#1{#1}\fi

\bibitem[{Alihosseini et~al.(2019)Alihosseini, Montahaei, and Soleymani~Baghshah}]{FBD}
Danial Alihosseini, Ehsan Montahaei, and Mahdieh Soleymani~Baghshah. 2019.
\newblock Jointly measuring diversity and quality in text generation models.
\newblock In {\em Proceedings of the Workshop on Methods for Optimizing and Evaluating Neural Language Generation\/}. Association for Computational Linguistics, Minneapolis, Minnesota, pages 90--98.

\bibitem[{Anderson et~al.(2016)Anderson, Fernando, Johnson, and Gould}]{spice}
Peter Anderson, Basura Fernando, Mark Johnson, and Stephen Gould. 2016.
\newblock Spice: Semantic propositional image caption evaluation.
\newblock In {\em Computer Vision--ECCV 2016: 14th European Conference, Amsterdam, The Netherlands, October 11-14, 2016, Proceedings, Part V 14\/}. Springer, pages 382--398.

\bibitem[{Brown et~al.(2020)Brown, Mann, Ryder, Subbiah, Kaplan, Dhariwal, Neelakantan, Shyam, Sastry, Askell et~al.}]{GPT3:2020}
Tom Brown, Benjamin Mann, Nick Ryder, Melanie Subbiah, Jared~D Kaplan, Prafulla Dhariwal, Arvind Neelakantan, Pranav Shyam, Girish Sastry, Amanda Askell, et~al. 2020.
\newblock Language models are few-shot learners.
\newblock {\em Advances in neural information processing systems\/} 33:1877--1901.

\bibitem[{Chen et~al.(2023)Chen, Wang, Jiang, Shi, and Xu}]{chen:2023:summarization}
Yi~Chen, Rui Wang, Haiyun Jiang, Shuming Shi, and Ruifeng Xu. 2023.
\newblock Exploring the use of large language models for reference-free text quality evaluation: A preliminary empirical study.
\newblock {\em arXiv preprint arXiv:2304.00723\/} .

\bibitem[{Davis and Marcus(2015)}]{Davis_2015}
Ernest Davis and Gary Marcus. 2015.
\newblock Commonsense reasoning and commonsense knowledge in artificial intelligence.
\newblock {\em Communications of the ACM\/} 58(9):92--103.

\bibitem[{Devlin et~al.(2019)Devlin, Chang, Lee, and Toutanova}]{BERT}
Jacob Devlin, Ming-Wei Chang, Kenton Lee, and Kristina Toutanova. 2019.
\newblock {BERT}: Pre-training of deep bidirectional transformers for language understanding.
\newblock In {\em Proceedings of the 2019 Conference of the North {A}merican Chapter of the Association for Computational Linguistics: Human Language Technologies, Volume 1 (Long and Short Papers)\/}. Association for Computational Linguistics, Minneapolis, Minnesota, pages 4171--4186.

\bibitem[{Dong et~al.(2022)Dong, Li, Dai, Zheng, Wu, Chang, Sun, Xu, and Sui}]{dong:2022:icl}
Qingxiu Dong, Lei Li, Damai Dai, Ce~Zheng, Zhiyong Wu, Baobao Chang, Xu~Sun, Jingjing Xu, and Zhifang Sui. 2022.
\newblock A survey for in-context learning.
\newblock {\em arXiv preprint arXiv:2301.00234\/} .

\bibitem[{Fan et~al.(2018)Fan, Lewis, and Dauphin}]{fan:2018:hierarchical}
Angela Fan, Mike Lewis, and Yann Dauphin. 2018.
\newblock Hierarchical neural story generation.
\newblock In {\em Proceedings of the 56th Annual Meeting of the Association for Computational Linguistics (Volume 1: Long Papers)\/}. Association for Computational Linguistics.

\bibitem[{Fan et~al.(2020)Fan, Gong, Wei, Wang, Huang, Jiao, Huang, Duan, and Zhang}]{Ekibart:2020}
Zhihao Fan, Yeyun Gong, Zhongyu Wei, Siyuan Wang, Yameng Huang, Jian Jiao, Xuan-Jing Huang, Nan Duan, and Ruofei Zhang. 2020.
\newblock An enhanced knowledge injection model for commonsense generation.
\newblock In {\em Proceedings of the 28th International Conference on Computational Linguistics\/}. pages 2014--2025.

\bibitem[{Gao et~al.(2021)Gao, Yao, and Chen}]{gao:2021:simcse}
Tianyu Gao, Xingcheng Yao, and Danqi Chen. 2021.
\newblock Simcse: Simple contrastive learning of sentence embeddings.
\newblock In {\em Proceedings of the 2021 Conference on Empirical Methods in Natural Language Processing\/}. pages 6894--6910.

\bibitem[{Gupta et~al.(2018)Gupta, Agarwal, Singh, and Rai}]{gupta:2018:deep}
Ankush Gupta, Arvind Agarwal, Prawaan Singh, and Piyush Rai. 2018.
\newblock A deep generative framework for paraphrase generation.
\newblock In {\em Proceedings of the aaai conference on artificial intelligence\/}. volume 32 (1).

\bibitem[{Heusel et~al.(2017)Heusel, Ramsauer, Unterthiner, Nessler, and Hochreiter}]{FID}
Martin Heusel, Hubert Ramsauer, Thomas Unterthiner, Bernhard Nessler, and Sepp Hochreiter. 2017.
\newblock Gans trained by a two time-scale update rule converge to a local nash equilibrium.
\newblock In I.~Guyon, U.~Von Luxburg, S.~Bengio, H.~Wallach, R.~Fergus, S.~Vishwanathan, and R.~Garnett, editors, {\em Advances in Neural Information Processing Systems\/}. Curran Associates, Inc., volume~30.

\bibitem[{Holtzman et~al.(2019)Holtzman, Buys, Du, Forbes, and Choi}]{holtzman:2019:sample}
Ari Holtzman, Jan Buys, Li~Du, Maxwell Forbes, and Yejin Choi. 2019.
\newblock The curious case of neural text degeneration.
\newblock {\em arXiv preprint arXiv:1904.09751\/} .

\bibitem[{Hu et~al.(2022)Hu, yelong shen, Wallis, Allen-Zhu, Li, Wang, Wang, and Chen}]{LORA}
Edward~J Hu, yelong shen, Phillip Wallis, Zeyuan Allen-Zhu, Yuanzhi Li, Shean Wang, Lu~Wang, and Weizhu Chen. 2022.
\newblock Lo{RA}: Low-rank adaptation of large language models.
\newblock In {\em International Conference on Learning Representations\/}.

\bibitem[{Hwang et~al.(2023)Hwang, Thost, Shwartz, and Ma}]{hwang:2023:knowledge}
EunJeong Hwang, Veronika Thost, Vered Shwartz, and Tengfei Ma. 2023.
\newblock \href{https://aclanthology.org/2023.emnlp-main.37}{Knowledge graph compression enhances diverse commonsense generation}.
\newblock In Houda Bouamor, Juan Pino, and Kalika Bali, editors, {\em Proceedings of the 2023 Conference on Empirical Methods in Natural Language Processing\/}. Association for Computational Linguistics, Singapore, pages 558--572.
\newblock \href{https://aclanthology.org/2023.emnlp-main.37}{https://aclanthology.org/2023.emnlp-main.37}.

\bibitem[{Kingma and Ba(2015)}]{ADAM}
Diederik~P. Kingma and Jimmy~Lei Ba. 2015.
\newblock Adam: A method for stochastic optimization.
\newblock In {\em Proc. of ICLR\/}.

\bibitem[{Lewis et~al.(2020)Lewis, Liu, Goyal, Ghazvininejad, Mohamed, Levy, Stoyanov, and Zettlemoyer}]{BART:2019}
Mike Lewis, Yinhan Liu, Naman Goyal, Marjan Ghazvininejad, Abdelrahman Mohamed, Omer Levy, Veselin Stoyanov, and Luke Zettlemoyer. 2020.
\newblock Bart: Denoising sequence-to-sequence pre-training for natural language generation, translation, and comprehension.
\newblock In {\em Proceedings of the 58th Annual Meeting of the Association for Computational Linguistics\/}. pages 7871--7880.

\bibitem[{Li et~al.(2023)Li, Wang, Guo, Song, Tan, Hassan, Menezes, Xiao, Bian, and Zhu}]{li:2023:deliberate}
Bei Li, Rui Wang, Junliang Guo, Kaitao Song, Xu~Tan, Hany Hassan, Arul Menezes, Tong Xiao, Jiang Bian, and JingBo Zhu. 2023.
\newblock Deliberate then generate: Enhanced prompting framework for text generation.

\bibitem[{Li et~al.(2021)Li, Gong, Jiao, Zhang, Baldwin, and Duan}]{KFCNet:2021}
Haonan Li, Yeyun Gong, Jian Jiao, Ruofei Zhang, Timothy Baldwin, and Nan Duan. 2021.
\newblock Kfcnet: Knowledge filtering and contrastive learning for generative commonsense reasoning.
\newblock In {\em Findings of the Association for Computational Linguistics: EMNLP 2021\/}. pages 2918--2928.

\bibitem[{Li et~al.(2016)Li, Galley, Brockett, Gao, and Dolan}]{li:2016:distinct}
Jiwei Li, Michel Galley, Chris Brockett, Jianfeng Gao, and William~B Dolan. 2016.
\newblock A diversity-promoting objective function for neural conversation models.
\newblock In {\em Proceedings of the 2016 Conference of the North American Chapter of the Association for Computational Linguistics: Human Language Technologies\/}. pages 110--119.

\bibitem[{Li et~al.(2018)Li, Ding, and Liu}]{li:2018:story}
Zhongyang Li, Xiao Ding, and Ting Liu. 2018.
\newblock Generating reasonable and diversified story ending using sequence to sequence model with adversarial training.
\newblock In {\em Proceedings of the 27th International Conference on Computational Linguistics\/}. pages 1033--1043.

\bibitem[{Lin et~al.(2020)Lin, Zhou, Shen, Zhou, Bhagavatula, Choi, and Ren}]{CommonGen}
Bill~Yuchen Lin, Wangchunshu Zhou, Ming Shen, Pei Zhou, Chandra Bhagavatula, Yejin Choi, and Xiang Ren. 2020.
\newblock {C}ommon{G}en: A constrained text generation challenge for generative commonsense reasoning.
\newblock In {\em Findings of the Association for Computational Linguistics: EMNLP 2020\/}. Association for Computational Linguistics, Online, pages 1823--1840.

\bibitem[{Liu et~al.(2023)Liu, Huang, Zhu, and Chang}]{liu:2023:dimongen}
Chenzhengyi Liu, Jie Huang, Kerui Zhu, and Kevin Chen-Chuan Chang. 2023.
\newblock \href{https://doi.org/10.18653/v1/2023.acl-long.260}{{D}imon{G}en: Diversified generative commonsense reasoning for explaining concept relationships}.
\newblock In Anna Rogers, Jordan Boyd-Graber, and Naoaki Okazaki, editors, {\em Proceedings of the 61st Annual Meeting of the Association for Computational Linguistics (Volume 1: Long Papers)\/}. Association for Computational Linguistics, Toronto, Canada, pages 4719--4731.
\newblock \href{https://doi.org/10.18653/v1/2023.acl-long.260}{https://doi.org/10.18653/v1/2023.acl-long.260}.

\bibitem[{Liu et~al.(2021)Liu, Wan, He, Peng, and Yu}]{kgbart:2021}
Ye~Liu, Yao Wan, Lifang He, Hao Peng, and Philip~S Yu. 2021.
\newblock Kg-bart: Knowledge graph-augmented bart for generative commonsense reasoning.
\newblock In {\em Proceedings of the AAAI Conference on Artificial Intelligence\/}. volume~35, pages 6418--6425.

\bibitem[{OpenAI(2023{\natexlab{a}})}]{openai:2023:gpt4}
OpenAI. 2023{\natexlab{a}}.
\newblock Gpt-4 technical report.

\bibitem[{OpenAI(2023{\natexlab{b}})}]{openai:2023:chatgpt}
OpenAI. 2023{\natexlab{b}}.
\newblock Introducing chatgpt.
\newblock \url{https://openai.com/blog/chatgpt}.
\newblock Accessed: 2023-11-23.

\bibitem[{Papineni et~al.(2002)Papineni, Roukos, Ward, and Zhu}]{bleu}
Kishore Papineni, Salim Roukos, Todd Ward, and Wei-Jing Zhu. 2002.
\newblock Bleu: a method for automatic evaluation of machine translation.
\newblock In {\em Proceedings of the 40th annual meeting of the Association for Computational Linguistics\/}. pages 311--318.

\bibitem[{Peeperkorn et~al.(2024)Peeperkorn, Kouwenhoven, Brown, and Jordanous}]{peeperkorn:2024:temperature}
Max Peeperkorn, Tom Kouwenhoven, Dan Brown, and Anna Jordanous. 2024.
\newblock Is temperature the creativity parameter of large language models?
\newblock {\em arXiv preprint arXiv:2405.00492\/} .

\bibitem[{Shen et~al.(2019)Shen, Ott, Auli, and Ranzato}]{shen:2019:mixture}
Tianxiao Shen, Myle Ott, Michael Auli, and Marc’Aurelio Ranzato. 2019.
\newblock Mixture models for diverse machine translation: Tricks of the trade.
\newblock In {\em International conference on machine learning\/}. PMLR, pages 5719--5728.

\bibitem[{Speer et~al.(2017)Speer, Chin, and Havasi}]{speer:2017:conceptnet}
Robyn Speer, Joshua Chin, and Catherine Havasi. 2017.
\newblock Conceptnet 5.5: An open multilingual graph of general knowledge.
\newblock In {\em Proceedings of the AAAI conference on artificial intelligence\/}. volume 31 (1).

\bibitem[{Sutskever et~al.(2014)Sutskever, Vinyals, and Le}]{sutskever:2014:sequence}
Ilya Sutskever, Oriol Vinyals, and Quoc~V Le. 2014.
\newblock Sequence to sequence learning with neural networks.
\newblock {\em Advances in neural information processing systems\/} 27.

\bibitem[{Tevet and Berant(2021)}]{tevet:2021:evaluating}
Guy Tevet and Jonathan Berant. 2021.
\newblock Evaluating the evaluation of diversity in natural language generation.
\newblock In {\em Proceedings of the 16th Conference of the European Chapter of the Association for Computational Linguistics: Main Volume\/}. pages 326--346.

\bibitem[{Touvron et~al.(2023)Touvron, Martin, Stone, Albert, Almahairi, Babaei, Bashlykov, Batra, Bhargava, Bhosale et~al.}]{touvron:2023:llama}
Hugo Touvron, Louis Martin, Kevin Stone, Peter Albert, Amjad Almahairi, Yasmine Babaei, Nikolay Bashlykov, Soumya Batra, Prajjwal Bhargava, Shruti Bhosale, et~al. 2023.
\newblock Llama 2: Open foundation and fine-tuned chat models.
\newblock {\em arXiv preprint arXiv:2307.09288\/} .

\bibitem[{Wang et~al.(2020)Wang, Liang, Jin, Wang, Zhu, and Zhang}]{wang:2020:semeval}
Cunxiang Wang, Shuailong Liang, Yili Jin, Yilong Wang, Xiaodan Zhu, and Yue Zhang. 2020.
\newblock Semeval-2020 task 4: Commonsense validation and explanation.
\newblock In {\em Proceedings of the Fourteenth Workshop on Semantic Evaluation\/}. pages 307--321.

\bibitem[{Wang et~al.(2023)Wang, Li, Dai, Chen, Zhou, Meng, Zhou, and Sun}]{wang-etal-2023-label}
Lean Wang, Lei Li, Damai Dai, Deli Chen, Hao Zhou, Fandong Meng, Jie Zhou, and Xu~Sun. 2023.
\newblock Label words are anchors: An information flow perspective for understanding in-context learning.
\newblock In Houda Bouamor, Juan Pino, and Kalika Bali, editors, {\em Proceedings of the 2023 Conference on Empirical Methods in Natural Language Processing\/}. Association for Computational Linguistics, Singapore, pages 9840--9855.

\bibitem[{Yu et~al.(2022)Yu, Zhu, Qin, Zhang, Zhao, and Jiang}]{yu:2022:diversifying}
Wenhao Yu, Chenguang Zhu, Lianhui Qin, Zhihan Zhang, Tong Zhao, and Meng Jiang. 2022.
\newblock Diversifying content generation for commonsense reasoning with mixture of knowledge graph experts.
\newblock In {\em Proceedings of the 2nd Workshop on Deep Learning on Graphs for Natural Language Processing (DLG4NLP 2022)\/}. pages 1--11.

\bibitem[{Zhang et~al.(2023)Zhang, Bollegala, and Peng}]{zhang-etal-2023-learning-predict}
Tianhui Zhang, Danushka Bollegala, and Bei Peng. 2023.
\newblock Learning to predict concept ordering for common sense generation.
\newblock In Jong~C. Park, Yuki Arase, Baotian Hu, Wei Lu, Derry Wijaya, Ayu Purwarianti, and Adila~Alfa Krisnadhi, editors, {\em Proceedings of the 13th International Joint Conference on Natural Language Processing and the 3rd Conference of the Asia-Pacific Chapter of the Association for Computational Linguistics (Volume 2: Short Papers)\/}. Association for Computational Linguistics, Nusa Dua, Bali, pages 10--19.

\bibitem[{Zhang et~al.(2020)Zhang, Kishore, Wu, Weinberger, and Artzi}]{zhang:2020:bertscore}
Tianyi Zhang, Varsha Kishore, Felix Wu, Kilian~Q. Weinberger, and Yoav Artzi. 2020.
\newblock Bertscore: Evaluating text generation with bert.

\bibitem[{Zheng et~al.(2023)Zheng, Chiang, Sheng, Zhuang, Wu, Zhuang, Lin, Li, Li, Xing et~al.}]{zheng:2023:judging}
Lianmin Zheng, Wei-Lin Chiang, Ying Sheng, Siyuan Zhuang, Zhanghao Wu, Yonghao Zhuang, Zi~Lin, Zhuohan Li, Dacheng Li, Eric Xing, et~al. 2023.
\newblock Judging llm-as-a-judge with mt-bench and chatbot arena.
\newblock {\em arXiv preprint arXiv:2306.05685\/} .

\bibitem[{Zhu et~al.(2018)Zhu, Lu, Zheng, Guo, Zhang, Wang, and Yu}]{zhu:2018:selfbleu}
Yaoming Zhu, Sidi Lu, Lei Zheng, Jiaxian Guo, Weinan Zhang, Jun Wang, and Yong Yu. 2018.
\newblock Texygen: A benchmarking platform for text generation models.
\newblock In {\em The 41st international ACM SIGIR conference on research \& development in information retrieval\/}. pages 1097--1100.

\end{thebibliography}
\bibliographystyle{acl_natbib}

\appendix
\section*{Supplementary Appendix}
\section{Mixture of Experts}
\label{sec:moe}
Given an input $x$, its corresponding LLM-generated sentences are divided into three random subsets. For each subset $G_i = \{g^i_1,...,g^i_k\}$, alongside the input $x$, we concatenate their token sequences with a separate latent variable $z_i$, resulting in the final input $x_i^{f}$. The $z_i$ is a randomly initialised sequence of tokens.
\begin{equation}
    x_i^{f} = z_i[CLS] x [SEP] g^i_1 [SEP]...g^i_k
\end{equation}
We train the model using the hard Expectation-Maximization (EM) approach, where during the E-step, for each $x_i^{f}$ and its corresponding target $y_i^{target}$, we identify the input that yields the highest probability as the best training example, with $\theta$ representing the generator model parameters: 

The model is trained using hard-EM by assigning full responsibility to the expert with the largest joint probability. In the E-step, for each input $x_i^{fin}$ and target $y_i^{tgt}$, choose the best input with the highest probability, to construct the training examples, where $\theta$ is the model's parameters.
\begin{equation}
    y_i^{tgt} = \underset{y_j}{\mathrm{argmax}} \, p(y_j|x_i^{f};\theta)
\end{equation}
Subsequently, in the M-step, we use these selected training examples to fine-tune the generator models. During inference, we input all diversified, context-aware inputs into the generator model to yield a range of diverse outputs.

\begin{table*}[t]
\centering
\resizebox{\textwidth}{!}{%
\begin{tabular}{@{}lrrrrrrrrrrrrr@{}}
\toprule
                               & & \multicolumn{2}{c}{Semantic Diversity $\Downarrow$} & \multicolumn{2}{c}{Corpus Diversity $\Uparrow$} & \multicolumn{2}{c}{Pairwise Diversity $\Downarrow$} & \multicolumn{4}{c}{Quality $\Uparrow$}                                & \multicolumn{2}{c}{Combined} \\ 
                               \cmidrule(r){3-4} \cmidrule(lr){5-6} \cmidrule(lr){7-8} \cmidrule(lr){9-12} \cmidrule(l){13-14}
                               & & self-cosSim & self-BERTScore & Entropy-4 & Distinct-4 & self-BLEU-3 & self-BLEU-4 & BLEU-3 & BLEU-4 & SPICE & BERTScore & Harmonic $\Uparrow$& FBD $\Downarrow$\\ 
\midrule
\midrule
        \multirow{4}{*}{$T=0$} & \textbf{Fine-tune} & 100.0 & 100.0 & 9.15 & 14.1 & 100.0 & 100.0 & 45.6 & \textit{34.9} & 34.4 & \textit{71.3} & 0.0 & 69.7 \\ 

 & \textbf{Default} & 100.0 & 100.0 & 9.12 & 15 & 100.0 & 100.0 & 40.8 & 30.4 & 28.5 & 67.6 & 0.0 & 69.7 \\ 
 & \textbf{Diversified} & 86.7 & 74 & 10.8 & 77.8 & 52.2 & 43.4 & 46.2 & 36.4 & 28.6 & 66.6 & 61.2 & 54.9 \\ 
 & \textbf{\ac{Prop}} & \textit{\textbf{86}} & \textit{\textbf{72.2}} & \textit{\textbf{10.9}} & \textit{\textbf{80}} & \textbf{\textit{46.5}} & \textit{\textbf{37.6}} & \textit{\textbf{48.1}} & \textit{\textbf{38.3}} & \textbf{29.1} & \textbf{67.7} & \textit{\textbf{65}} & \textit{\textbf{53.5}} \\ 
 \midrule
 \midrule
\multirow{4}{*}{$T=0.5$} & \textbf{Fine-tune} & \textit{83.6} & 81.6 & 10.6 & 65.4 & 63.8 & 55.7 & \textit{56.3} & \textit{46.8} & \textit{36.1} & \textit{73.8} & 55.4 & 58.3 \\ 
 & \textbf{Default} & 96.1 & 94.9 & 9.75 & 36.9 & 88.9 & 86.8 & \textbf{47.6} & \textbf{37.3} & \textbf{29.5} & \textbf{69.6} & 22.4 & 63.7 \\ 
 & \textbf{Diversified} & 86.4 & 73 & 10.9 & 79.8 & 49.9 & 40.8 & 46 & 36.5 & 28.6 & 66.5 & 62.6 & 55.4 \\ 
 & \textbf{\ac{Prop}} & \textbf{85.1} & \textit{\textbf{70}} & \textit{\textbf{10.9}} & \textit{\textbf{84.2}} & \textit{\textbf{39.4}} & \textit{\textbf{29.5}} & 48.6 & 39.1 & 29.2 & 68.1 & \textbf{\textit{69.3}} & \textbf{\textit{53.4}} \\ 
\midrule
\midrule
\multirow{4}{*}{$T=1$} & \textbf{Fine-tune}                       & \textit{64.7} & \textit{55.9}  & \textit{11.4}      & \textit{91.1}       & 26.9 & 17.9 & 41.2   & 32.1   & 30.3  & 64.2      & 72.1          & 51.9 \\
& \textbf{\default}            & 93.3          & 88.7           & 10.2      & 53.7       & 77.2 & 72.4 & \textit{\textbf{50.8}}   & \textit{\textbf{40.9}}   & \textit{\textbf{30.1}}  & \textit{\textbf{70.4}}      & 39.6          & 60.2 \\
& \textbf{\diversified}           & 85.2          & 69.8           & \textbf{11.0}      & 83.7       & 44.4 & 34.9 & 44.3   & 34.6   & 28.5  & 65.0      & 65.4          & 53.9 \\
& \textbf{\ac{Prop}}            & \textbf{83.5}          & \textbf{66.2}           & \textbf{11.0} & \textbf{88.5} & \textit{\textbf{31.0}} & \textit{\textbf{21.0}} & 47.4   & 37.7   & 29.1  & 67.4      & \textit{\textbf{72.7}}          & \textit{\textbf{51.8}} \\
 \midrule
 \midrule
\multirow{4}{*}{$T=1.5$} & \textbf{Fine-tune} & \textit{25.7} & \textit{0.0} & \textit{11.9} & \textit{100.0} & \textit{2.5} & \textit{1.8} & 8.1 & 4.3 & 11 & 16 & 27.5 & 67.8 \\ 
 & \textbf{Default} & 90.4 & 81.5 & 10.5 & 68.4 & 63.5 & 56.1 & \textbf{51.4} & \textbf{41.9} & \textbf{29.9} & \textbf{70.1} & 54 & 56.5 \\ 
 & \textbf{Diversified} & 67.7 & \textbf{59.3} & 11.2 & 89.5 & 30.6 & 22.3 & 39.3 & 29.7 & 26.9 & 61.9 & 68.9 & 54 \\ 
 & \textbf{\ac{Prop}} & \textbf{78.3} & 59.8 & \textbf{11.2} & \textbf{92.8} & \textbf{20.9} & \textbf{12.4} & 44.1 & 34.7 & 27.9 & 65.4 & \textbf{\textit{74.9}} & \textbf{\textit{51.6}} \\ 
\bottomrule
\end{tabular}}
\caption{Diversity and quality scores on four temperature settings ($T=0/0.5/1/1.5$) on the CommonGen dataset. The results show that our proposed method ICD performs well across different temperatures.
Best results for each metric at each temperature setting are shown in \textit{italics}, while the best performing \ac{ICL} results are shown in \textbf{bold}.
}
\label{tbl:temperature-res}
\end{table*}

\section{Impact of sampling temperature on the diversity and quality}
\label{sec:temperature}
In this section, we investigate the impact of temperature on various methods on the CommonGen dataset.  Although GPT-3.5 also provides the nucleus sampling beyond sampling temperature, we specifically focus on the general performance of \ac{Prop} under different temperature settings and set nucleus sampling hyper-parameter to 1. Our experiments are conducted on the \texttt{GPT-3.5-turbo-0613}. \autoref{tbl:temperature-res} demonstrates that \ac{Prop} consistently outperforms both~\default~and~\diversified~on the Combined metrics across all temperature settings, which aligns with our findings in~\autoref{sec:exp:GCR}. Moreover, \ac{Prop} exhibits less sensitivity to temperature variations compared to the other baselines and performs better on Combined metrics with the increase of temperature, which can be considered as an additional advantage of our proposed method. 

Furthermore, we observe that the \textbf{fine-tune} method is also significantly influenced by temperature sampling on the \ac{GCR} task. At $T=1.5$, the \default~baseline, which applies \ac{ICL} on the same base model GPT-3.5-turbo, outperforms the \textbf{fine-tune} method. The fine-tuned model generates responses that are of very low quality, consisting mostly of nonsensical word combinations.
For example, given the input ``sidewalk leash dog walk'', the fine-tune method would generate the random sequence: \textit{A owners with 2 kangaroos trying to walk their yappy circus bear disguised as a gers?'''texturesumm} while \ac{Prop} generates sentence than covers the task requirement: \textit{A dog walks on a sidewalk, attached to a leash.}
Therefore, we conclude that increasing temperature of the decoder is not a suitable strategy for improving diversity in GCR.

\begin{table*}[t]
\centering
\resizebox{\textwidth}{!}{%
\begin{tabular}{@{}lrrrrrrrrrrrr@{}}
\toprule
                               & \multicolumn{2}{c}{Semantic Diversity $\Downarrow$} & \multicolumn{2}{c}{Corpus Diversity $\Uparrow$} & \multicolumn{2}{c}{Pairwise Diversity $\Downarrow$} & \multicolumn{4}{c}{Quality $\Uparrow$}                                & \multicolumn{2}{c}{Combined} \\ 
                               \cmidrule(r){2-3} \cmidrule(lr){4-5} \cmidrule(lr){6-7} \cmidrule(lr){8-11} \cmidrule(l){12-13}
                               & self-cosSim & self-BERTScore & Entropy-4 & Distinct-4 & self-BLEU-3 & self-BLEU-4 & BLEU-3 & BLEU-4 & SPICE & BERTScore & Harmonic $\Uparrow$& FBD $\Downarrow$\\ 
\midrule
\multicolumn{13}{@{}c}{\textbf{CommonGen}} \\
\midrule
\textbf{Fine-tune}                       & 59.6 & 49.9 & \textit{11.4} & 93.3 & 22.8 & 14.5 & 35.8 & 26.8 & 27.6 & 59.1 & 69.9 & 52.4 \\
\midrule
\textbf{\default}            & 82.2 & 73.8 & 10.9 & 74.9 & 52.9 & 45.4 & \textbf{\textit{44.6}} & \textbf{\textit{34.9}} & \textbf{\textit{29.1}} & \textbf{\textit{67.1}} & 60.2 & 56.2 \\
\textbf{\diversified}           & \textbf{\textit{59.1}} & 53.3 & 11.3 & 91.3 & 23.6 & 16.4 & 32.6 & 23.7 & 24.3 & 58.2 & 68.6 & 53.2 \\
\textbf{\ac{Prop}}            & 59.3 & \textbf{\textit{49.8}} & \textbf{\textit{11.3}} & \textbf{\textit{93.7}} & \textbf{11.3} & \textbf{\textit{5.8}} & 34.2 & 24.9 & 25.5 & 60.1 & \textit{\textbf{73.4}} & \textit{\textbf{51.0}} \\
\midrule
\midrule
\multicolumn{13}{@{}c}{\textbf{ComVE}} \\
\midrule
\textbf{Fine-tune}                        & \textit{60.4} & 45.8 & 9.6 & \textit{93.8} & \textit{17.1} & 14.1 & \textit{27.9} & \textit{\textbf{19}} & 31.1 & \textit{52.3} & \textit{65.0} & \textit{47.3} \\
\midrule
\textbf{\default}            & 75.7 & 57.1 & 9.8 & 78.0 & 36.7 & 31.1 & \textbf{23.8} & \textbf{16.9} & \textit{\textbf{33}} & 49.2 & 57.4 & 60.8 \\
\textbf{\diversified}          & 64.7 & 42.3 & \textit{\textbf{10.0}} & 89.3 & 13.4 & 8.8 & 23.2 & 16.0 & 32.6 & \textbf{49.8} & 64.4 & \textbf{56.9} \\
\textbf{\ac{Prop}}           & \textbf{61.5} & \textit{\textbf{37.3}} & \textit{\textbf{10.0}} & \textbf{90.1} & \textit{\textbf{5.8}} & \textit{\textbf{3.0}} & 22.7 & 15.7 & 32.5 & 48.8 & \textbf{65.1} & 58.2 \\ 
\midrule
\midrule
\multicolumn{13}{@{}c}{\textbf{DimonGen}} \\
\midrule
\textbf{Fine-tune}                       & \textit{41} & \textit{29.5} & \textit{10.4} & \textit{99} & 5 & 2.2 & \textit{15.4} & 8.9 & \textit{14.6} & 39.4 & 56.2 & \textit{52.8} \\
\midrule
\textbf{\default}            & 64.0 & 48.6 & 10.3 & 95.0 & 17.9 & 13.1 & \textbf{13.6} & \textbf{7.9} & \textbf{14.4} & \textit{\textbf{41.3}} & 56 & 61.1 \\
\textbf{\diversified}           & 55.2 & 45.4 & 10.3 & 97 & 11.9 & 7.3 & 12.1 & 6.7 & 13.4 & 39.8 & 55.7 & 62 \\
\textbf{\ac{Prop}}            & \textbf{53.1} & \textbf{37.0} & \textbf{10.4} & \textbf{98} & \textit{\textbf{2.4}} & \textit{\textbf{0.9}} & 12.7 & 7.3 & 13.6 & 39 & \textit{\textbf{56.6}} & \textbf{61.1} \\
\bottomrule
\end{tabular}%
}
\caption{Performance on CommonGen, ComVE and DimonGen with \texttt{Vicuna-13b}. 
Best results on each task for each metric are shown in \textit{italics}, while the best performing \ac{ICL} results are shown in \textbf{bold}.
}
\label{tbl:full-vicuna}
\end{table*}

\section{Full results on Vicuna-13b model}
\label{sec:full-vicuna}
\autoref{tbl:full-vicuna} shows the full result on the open source Vicuna-13b model across three datasets. It reconfirms \ac{Prop}'s ability to balance both quality and diversity according to the combined metrics. Furthermore, we find that methods using the Vicuna model show lower quality than those using GPT-3.5-turbo while generating more diverse sentences.

\begin{figure*}[ht]
\centering
\begin{subfigure}[b]{0.9\linewidth}
    \includegraphics[width=\linewidth]{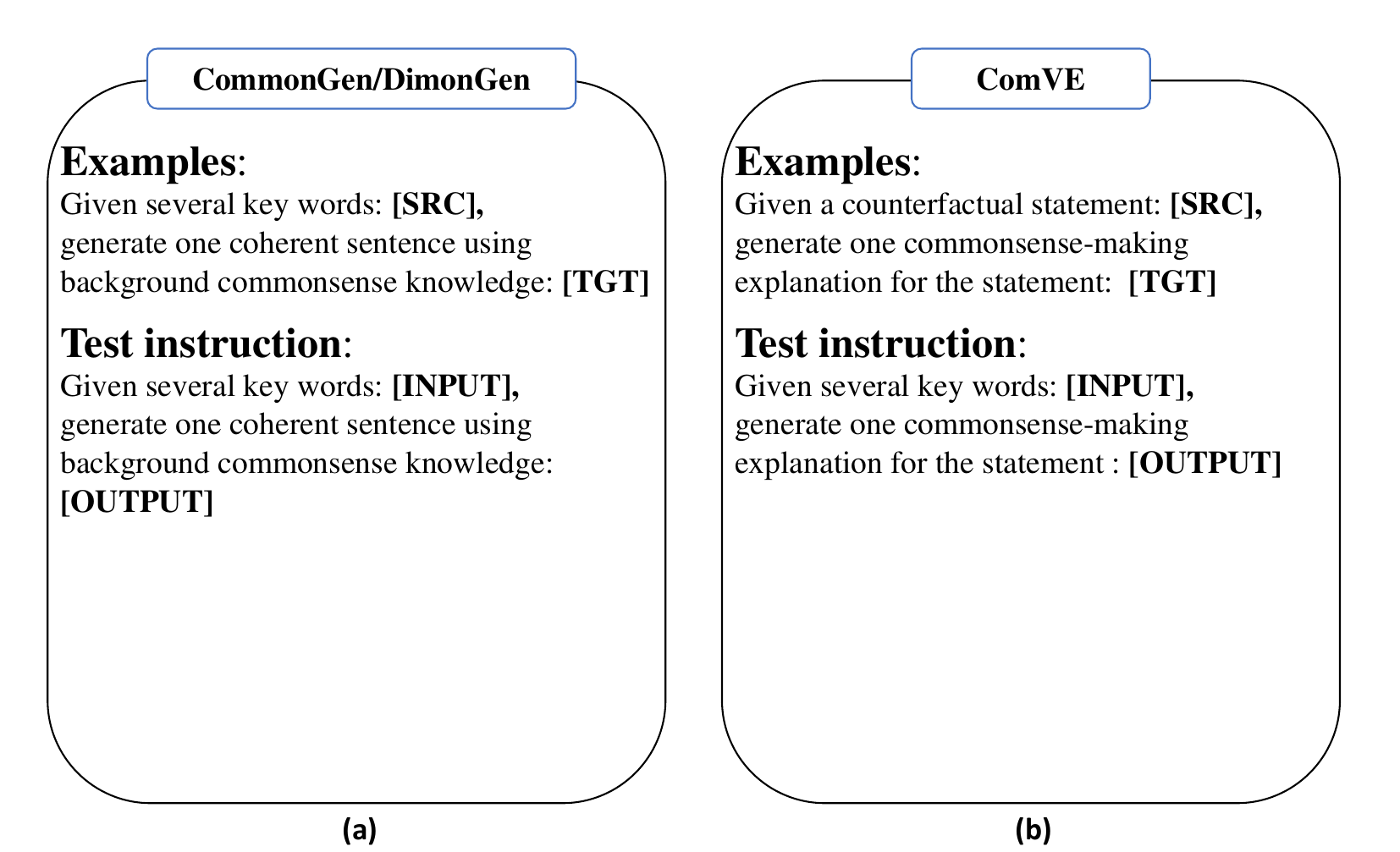}
    \caption{Default instructions}
    \label{fig:default-instructions}
\end{subfigure}
\hfill %
\begin{subfigure}[b]{0.9\linewidth}
    \includegraphics[width=\linewidth]{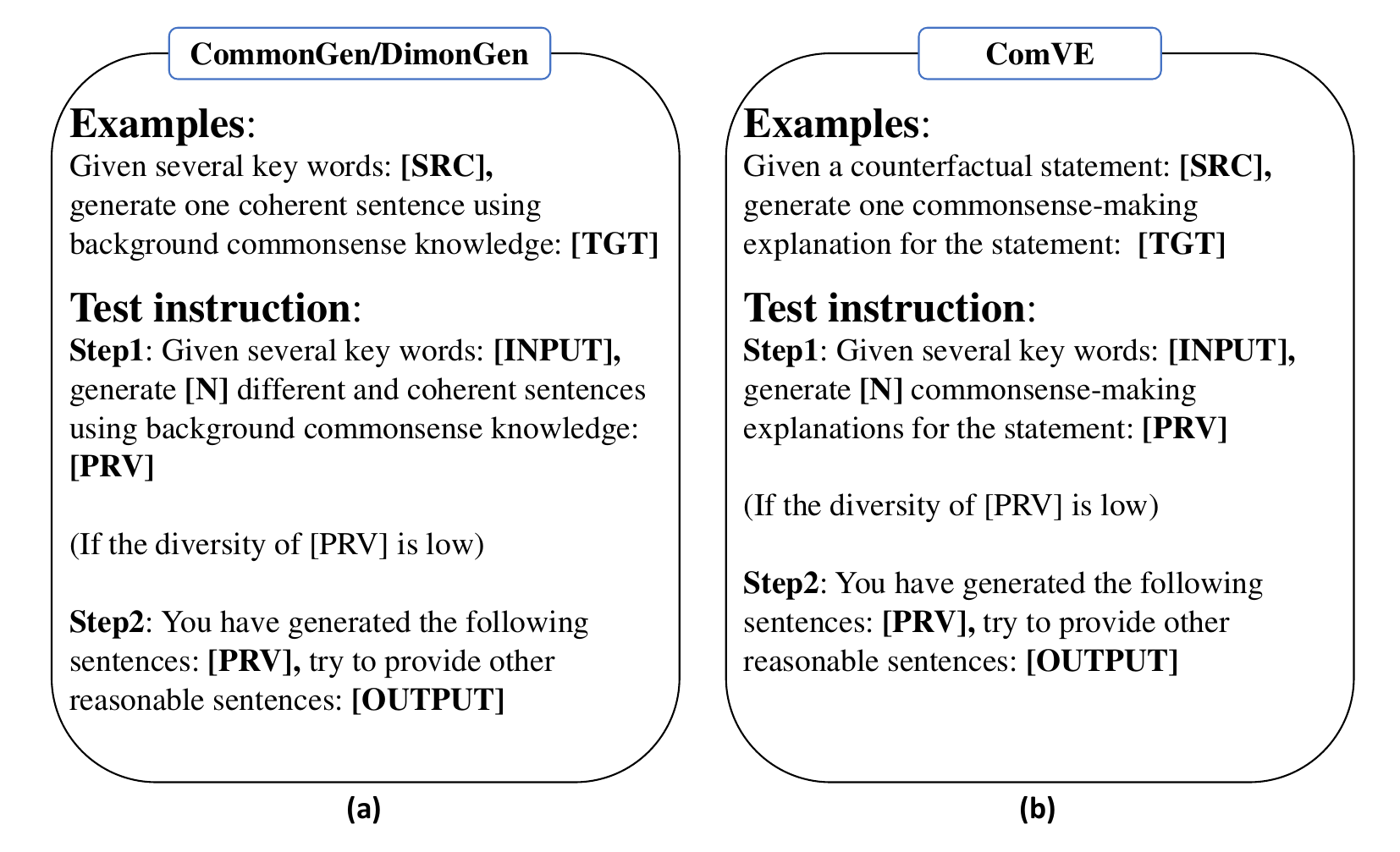}
    \caption{Diversified instructions}
    \label{fig:diversified-instructions}
\end{subfigure}
\caption{The templates used by the \default~and the \diversified~prompt instructions for the CommonGen/DimonGen (shown on the left, (a)) and ComVE (shown on the right, (b)) tasks.
Few-shot examples are included in each prompt where \textbf{[SRC]} denotes the set of input concepts and  \textbf{[TGT]} the corresponding sentences in CommonGen.
For a given set of \textbf{[INPUT]} concepts, the \ac{LLM} is then required to generate sentences at the slot \textbf{[OUTPUT]}.}
\label{fig:instructions}
\end{figure*}

\section{LLM Prompt Templates}
\label{sec:LLMTemplates}
~\autoref{fig:instructions} shows the templates that are used for the two \ac{GCR} tasks: CommonGen and ConVE. 
The \default~prompt is adapted from~\citet{li:2023:deliberate} and are task-specific.
On the other hand, the \diversified~prompt modifies the \default~prompt by appending a task-independent instruction that first checks whether the diversity of the sentences generated in Step 1 is low, and if presents the generated sentences to the \ac{LLM} and re-prompts it to generate more diverse set of sentences.

\begin{figure*}[t]
\centering
\includegraphics[width=1.1\linewidth]{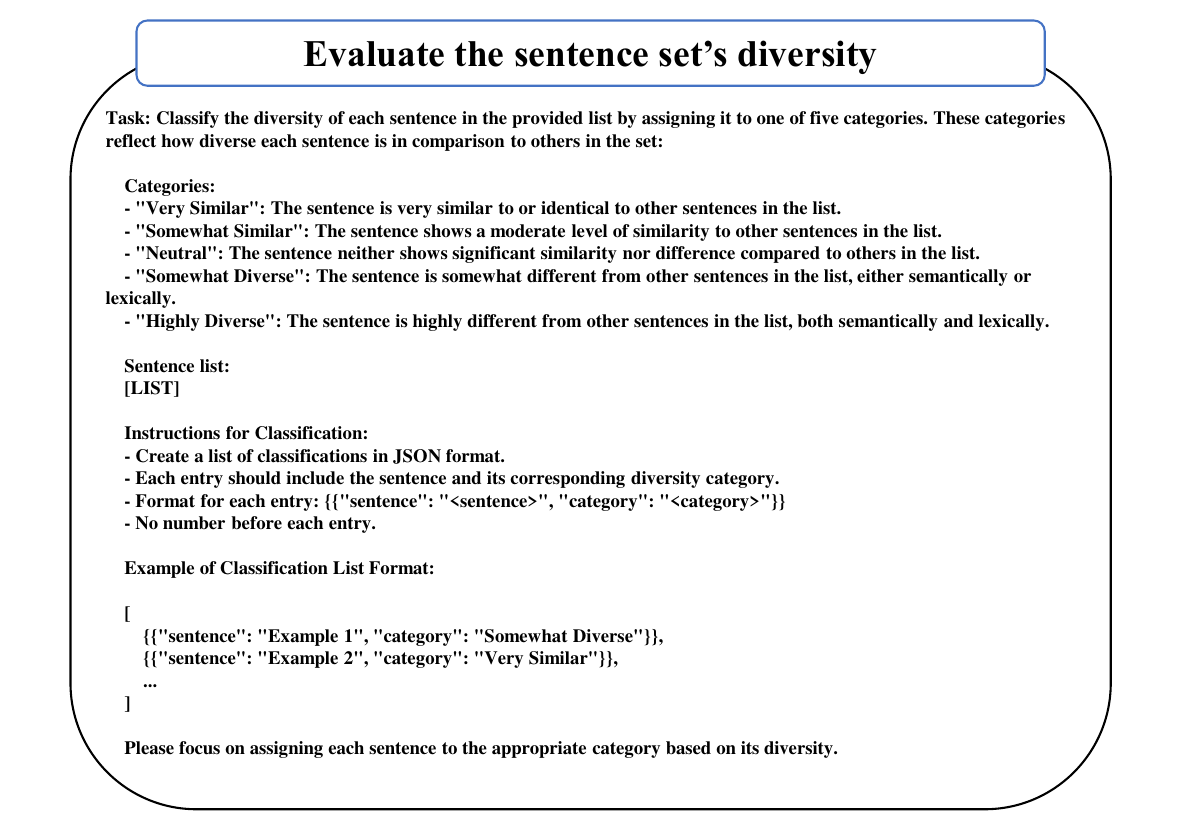}
\caption{The instructions provided to \texttt{GPT3.5-turbo} for predicting the diversity of a given set of sentences. Diversity is predicted according to five categories: \emph{very similar}, \emph{somewhat similar}, \emph{neutral}, \emph{somewhat diverse}, and \emph{highly diverse}.
Definitions of the categories are included within the instructions.
Next, the set of sentences to be evaluated for their diversity is presented.
Finally, the expected output format of the predictions is described at the end of the prompt.
 As recommended by~\citet{chen:2023:summarization}, we do not require the \ac{LLM} to provide  reasons for its predictions because it sometimes forces the model to focus on the reason generation than the prediction. }
\label{fig:diversitygpt}
\end{figure*}

We use \texttt{GPT3.5-turbo} to predict the diversity of a given set of sentences using the prompt shown in \autoref{fig:diversitygpt}.
This prompt uses five diversity categories (i.e. \emph{very similar}, \emph{somewhat similar}, \emph{neutral}, \emph{somewhat diverse}, and \emph{highly diverse}) with increasing diversity with their definitions.
Next, the set of sentences to be evaluated for their diversity is presented.
Finally, the expected output format of the predictions is described at the end of the prompt.
 As recommended by~\citet{chen:2023:summarization}, we do not require the \ac{LLM} to provide  reasons for its predictions because it sometimes forces the model to focus on the reason generation than the prediction. 
 
 After the LLM's evaluation, the predictions are mapped to values from 1 to 5 where 1 being highly similar to 5 being highly diverse. For each sentences set, we take the average of \ac{LLM} predictions over three independent runs.

\section{Human Evaluation}
As human-annotators, we recruited eight graduate students from the department of computer science who specialise in NLP and are fluent speakers in English.
We provided the human annotators with the same set of instructions as we provided to the \ac{LLM}s.
Apart from the instruction in~\autoref{fig:diversitygpt}, we also instruct the human annotators to focus on the diversity, and not so much on their commonsense quality, which we evaluate separately using semi-automatic metrics by comparing against human-written reference sentences in the evaluation benchmarks.
Moreover, we informed the annotators that their evaluations would be used in a comparative analysis with the scores generated by an \ac{LLM}.

\section{Additional Generation Examples}
We show additional sentences generated by our proposed methods for the CommonGen, ComVE and DimonGen datasets in~\autoref{tab:more_exe}.

\begin{table*}
\centering
\resizebox{\textwidth}{!}{%
\begin{tabular}{@{}ll@{}}
\toprule
\textbf{CommonGen} &  \\ \midrule
Keyword & dog throw frisbee catch \\ \midrule
Sentences & The dog eagerly waits for the frisbee to be thrown and then leaps to catch it in mid-air. \\
 & The owner tosses the frisbee and the dog quickly reacts, catching it effortlessly. \\
 & The dog excitedly chases after the thrown frisbee and skillfully catches it in its mouth. \\
 & After the frisbee is thrown, the dog jumps up high and grabs it, showcasing its impressive catching skills. \\
 & The dog enthusiastically catches the frisbee, bringing it back to the owner for another throw. \\
 & A dog catches a frisbee that was thrown. \\ \midrule
Keyword &  roast stick marshmallow fire \\ \midrule
Sentences & A marshmallow is being roasted on a stick over a fire. \\
 & People roast marshmallows on a stick over a fire. \\
 & During a campfire, people often stick marshmallows on a stick and roast them over the fire. \\
 & The act of roasting marshmallows involves sticking them on a stick and holding them over a fire. \\
 & People gather around a fire, roasting marshmallows on sticks. \\
 & Sticking a marshmallow on a stick and roasting it over a fire is a common activity. \\ 
 \midrule
 \midrule
\textbf{ComVE} &  \\ \midrule
Statement & She parachuted into the grocery store from a plane. \\ \midrule
Explanation & Parachuting is not a common method of entering a grocery store. \\
 & Grocery stores do not have landing areas for parachutes. \\
 & Parachuting into a grocery store from a plane is highly unlikely and not a normal activity. \\ \midrule
Statement & You can drive after drinking. \\ \midrule
Explanation & Drinking alcohol impairs your ability to drive, so it is not recommended. \\
 & It is illegal and unsafe to drive after drinking. \\
 & Driving after drinking is dangerous and against the law. \\ 
\midrule
 \midrule
\textbf{DimonGen} &  \\ \midrule
Keyword & cloud mountain \\ \midrule
Sentences & As we climbed the mountain, the clouds grew thicker, obscuring our view. \\
 & The mountain stood tall and majestic, with clouds swirling around its peaks. \\
 & The mountain peak pierced through the clouds, reaching towards the sky. \\ \midrule
Keyword & race victory \\ \midrule
Sentences & The underdog pulled off an unexpected victory in the race, leaving the favorite trailing behind. \\
 & With a burst of speed and determination, the runner sprinted towards the finish line, securing a triumphant victory. \\
 & After a fierce race, the champion celebrated their victory with a crowd cheering and fireworks lighting up the sky. \\
 \bottomrule
\end{tabular}%
}
\caption{More examples produced by our proposed \ac{Prop} method on the CommonGen, ComVE and DimonGen datasets.}
\label{tab:more_exe}
\end{table*}

\end{document}